\documentclass[9.5pt,journal,compsoc]{IEEEtran}

\makeatletter
\def\normalsize{\@setfontsize{\normalsize}{9.5pt}{11.5pt}}
\normalsize
\makeatother

\usepackage[nocompress]{cite}

\ifCLASSOPTIONcompsoc
  \usepackage[nocompress]{cite}
\else
  \usepackage{cite}
\fi

\usepackage[normalem]{ulem}

\usepackage[english]{babel}
\usepackage{blindtext}
\usepackage{textcomp}
\usepackage{amsmath}
\usepackage{makecell}
\usepackage{graphicx}
\usepackage{multirow}
\usepackage{fancyhdr}
\usepackage{xspace}
\usepackage{dsfont}
\usepackage{pgfplots}
\usepackage{multirow}
\usepackage{float}
\usepackage{microtype}
\usepackage{subfig}
\usepackage{booktabs}
\usepackage{hyperref}

\usepackage{comment}

\pgfplotsset{compat = 1.9}
\usetikzlibrary{patterns}

\def \OurSystem{TAG\xspace}

\begin{document}

\title{Expediting Distributed DNN Training with Device Topology-Aware Graph Deployment}

\author{Shiwei~Zhang,
        Xiaodong~Yi,
        Lansong~Diao,
        Chuan~Wu,
        Siyu~Wang,
        and Wei~Lin}

\IEEEtitleabstractindextext{
\begin{abstract}
    This paper presents \OurSystem, an automatic system to derive optimized DNN training graph and its deployment onto any
    device topology, for expedited training in device- and topology- heterogeneous ML clusters. We novelly combine both the
    DNN computation graph and the device topology graph as input to a graph neural network (GNN), and join the GNN with a
    search-based method to quickly identify optimized distributed training strategies. To reduce communication in a
    heterogeneous cluster, we further explore a lossless gradient compression technique and solve a combinatorial
    optimization problem to automatically apply the technique for training time minimization. We evaluate \OurSystem with
    various representative DNN models and device topologies, showing that it can achieve up to 4.56x training speed-up as
    compared to existing schemes. \OurSystem can produce efficient deployment strategies for both unseen DNN models and
    unseen device topologies, without heavy fine-tuning.
\end{abstract}

\begin{IEEEkeywords}
Distributed Systems, Machine Learning
\end{IEEEkeywords}}

\maketitle

\IEEEdisplaynontitleabstractindextext
\IEEEpeerreviewmaketitle

\section{Introduction}\label{sec:introduction}

Deep learning (DL) has powered a wide range of applications in various areas including computer vision
\cite{resnet,inception}, natural language processing \cite{gpt3,palm}, recommendation systems \cite{wang2018billion},
etc. Recent deep neural network (DNN) models feature a large number of parameters (e.g. BERT \cite{bert} with more than
340M parameters) to achieve superior performance \cite{gpt3,bert}. Large-scale distributed training using tens or
hundreds of GPUs on a cluster of machines has been the norm for training these models.

State-of-the-art distributed training largely exploits a homogeneous cluster, e.g., training Bert using 8 NVIDIA V100
GPUs \cite{biobert}. Nonetheless, modern AI clouds often host a number of server types equipped with different devices
(e.g., A100, V100 and P100 GPUs). Only allocating the same type of GPUs/machines to each training job may well result in
scattered idling resources, e.g., 1 V100 GPU on one machine and 2 P100 GPUs on another. Often, the scattered resources
cannot be allocated to one job due to lack of efficient support for training on a heterogeneous cluster with existing DL
frameworks, and hence largely remain idle. To better utilize the expensive AI infrastructure, enabling efficient
distributed training over heterogeneous devices is the key. 

Further, inter-connectivity and bandwidth across devices in an AI cloud often differ, due to different link types (e.g.,
PCIE or NVLink inside a machine), different co-location levels (e.g., machines in the same rack or not), etc. This adds
onto the heterogeneity of the machine learning (ML) cluster used to train a DNN model. 

Multiple related decisions are involved for deploying a DNN model onto heterogeneous, scattered resources for most
expedited training: On which device of which machine shall we place each operator (operation placement
\cite{placeto,gdp})? Shall we replicate one or a group of operators on multiple devices for data-parallel training
(operation replication \cite{heterog,autosync})? Should we use AllReduce \cite{horovod,nccl} or the parameter server
(PS) architecture \cite{ps,byteps} for parameter synchronization among replicated operators, and should gradients
produced by some operators be compressed to reduce inter-device communication (gradient compression
\cite{dgc,autosync})? These decisions jointly form an exponentially large strategy space. Current practice often falls
back to heuristics that consider one aspect of the strategy space at a time \cite{parallax,hdp}, resulting in less
efficient or even infeasible solutions.

Pioneering works on deploying DNN models onto heterogeneous computation resources adopt reinforcement learning and neural
networks for finding distributed training strategies \cite{grl,hdp,heterog}. However, their models do not generalize
to different device topologies and require training from scratch for each new resource configuration. This makes them
impractical for AI clouds, where new resource configurations are made for each job. A generic method that
can quickly find distributed training strategies for unseen device topologies is yet to be explored.

We present \OurSystem\footnote{We plan to open-source \OurSystem.}, an automatic DNN deployment framework that
efficiently produces optimized distributed training strategies for a given DNN model on heterogeneous resources.
\OurSystem exploits a heterogeneous graph neural network (GNN) \cite{heterognn} jointly with a search-based method to
make fine-grained decisions on operation replication, placement, parameter synchronization and gradient compression.

The key contributions of \OurSystem are summarized as follows:

$\triangleright$ An automatic DNN deployment framework is proposed that produces optimized training graph with
operation-level replication, for expedited training over any given device set and inter-device topology. It
automatically inserts necessary operations to ensure mathematical equivalence before and after modifying the original
DNN computation graph.

$\triangleright$ A heterogeneous GNN is designed which takes both computation graph and device topology as input, and
learns a generalizable policy to guide a Monte Carlo tree search (MCTS) \cite{mcts,puct} for efficient operation
placement strategies.

$\triangleright$ To further reduce communication overhead, we adopt sufficient factor broadcasting (SFB) \cite{sfb}, a
lossless gradient compression technique, and formulate a graph-cut problem to automatically decide subgraph duplication
and SFB's application in the training graph.

$\triangleright$ Extensive experiments are conducted over representative DNN models and various device topologies.
\OurSystem achieves up-to 4.56x speed-up as compared to data parallelism using NCCL AllReduce and state-of-the-art
training schemes on heterogeneous clusters. It can generate efficient deployment strategies for unseen DNN models on
unseen device topologies without heavy fine-tuning.

\section{Background}\label{sec:background}

\subsection{Distributed DNN training\label{sec:background_dist}}

DNN models implemented in modern ML frameworks, e.g., TensorFlow \cite{tensorflow}, Pytorch \cite{pytorch}, and MXNet
\cite{mxnet}, can be represented by directed acyclic graphs (DAG), referred to as the {\em computation graph} of the
respective DNN models. In a computation graph, each node is an operation (op) and the edges connecting the ops represent
tensors. The ops are placed and executed on computation devices (e.g., GPUs). If an op that produces a tensor and the op
that consumes the tensor are placed on different devices, the tensor needs to be transferred across devices.

Training a DNN is an iterative process. In each iteration, forward computation is performed on a batch of training data,
followed by backward propagation that calculates the gradients and updates the model parameters. Data parallelism (DP)
and model parallelism (MP) are two main paradigms for distributed training \cite{bert, mp}. With classic DP, each device
holds a full copy of the model and processes a batch of data independently.
MP puts different parts of the DNN model onto different devices, and
intermediate tensors are passed between the devices during training.

A number of studies have explored device placement, op replication and hybrid DP/MP for DNN training. GDP \cite{gdp}, GO
\cite{go}, PlaceTo \cite{placeto} and HeteroG \cite{heterog} use GNN to extract computation graph information and
generate device assignment of ops. HDP \cite{hdp} and Spotlight \cite{spotlight} use reinforcement learning (RL) to
train an LSTM for placement decisions. FlexFlow \cite{flexflow} uses the Markov Chain Monte Carlo (MCMC) search
algorithm to search for op placements, achieving hybrid parallelism. REGAL \cite{regal} uses RL and genetic algorithm
(GA) to co-optimize placement and scheduling. Pesto \cite{pesto} models placement and scheduling into
an integer linear program (ILP) and solves it with off-the-shelf ILP solvers.
None of the learning-based systems supports unseen device topologies, and retraining of the
respective GNN or LSTM models is required when the device topology changes.

\subsection{Shared AI infrastructure}

A state-of-the-art AI cloud typically includes multiple racks of servers, equipped with a few representative types of
GPUs of different generations. A common practice in production AI clouds is to allocate GPUs of the same type to one
training job, ideally located on the same machine or machines close to each other \cite{hived}. Such resource allocation
often leads to resource fragmentation, e.g., 1-2 unallocated GPUs scattered on different machines, causing substantial
wastage of expensive AI resources. Further, training jobs requesting a large number of GPUs of the same model
suffer from long queuing times \cite{mlaas}, as they have to wait for all resources to be available.

The existing DL frameworks (TensorFlow, PyTorch, MXNet, etc.) and training strategies mostly support efficient
distributed training over homogeneous clusters, and their training efficiency deteriorates substantially on
heterogeneous resources (i.e., GPUs with different computation and memory capacities, varying inter-device bandwidth).
A design to efficiently utilize the scattered, heterogeneous resources is of strong interest.

\subsection{Sufficient Factors Broadcasting}

Sufficient factor broadcasting utilizes the low-rank structures in gradient tensors to make mathematically equivalent
transformation on the computation graph \cite{sfb}. Sufficient factors are small tensors that can generate a gradient
tensor, usually by an outer product. In SFB, sufficient factors are broadcast to op replicas instead of the full
gradient tensor, potentially reducing communication time. SFB does not impose precision loss nor affect training
convergence. 

Chilimbi et al.~\cite{chilimbi2014project} exploit low-rank structures in the last layers of CNN and advocate explicitly
sending only the activation and error gradient vectors to the PSs. Poseidon \cite{zhang2017poseidon} broadcasts
sufficient factors for synchronizing gradients among workers.
They only support \texttt{MatMul} layers, limiting their usage on recent models with new types of ops.
These methods are not automatically enabled and an ML
developer needs to choose where to apply these optimizations for the best efficiency.

\section{Motivation and Challenges\label{sec:motivation}}

\subsection{Opportunities}

\textbf{Partial Data Parallelism.}
Existing frameworks either replicate an op on all GPUs, or place it on just one GPU in pure DP, MP, or hybrid DP/MP
\cite{grl,heterog,hdp,spotlight,placeto,gdp}. In a given GPU cluster, it is possible to replicate some ops on a subset
of nearby GPUs, but not on other GPUs, achieving a good trade-off between GPU utilization and parameter synchronization
overhead \cite{flexflow}.

\textbf{Topology-Aware Automatic Device Placement.}
In existing designs, the decision NN needs to be retrained when the device topology changes (e.g., devices to use or the
link bandwidth between devices change), because they do not take device topology as input to their NNs and the structure
of their NNs may also need to be altered when device topology changes. For example, the output dimension of the GNN in
HeteroG \cite{heterog} is $M+4$, where $M$ is the number of devices; when the number of devices changes, its GNN needs
to be retrained with a new output dimension. Designing a strategy making framework that can handle various device
topologies, as well as different DNNs, is a more general and practical solution. 

\textbf{Automatic Sufficient Factors Broadcasting.}
With the advance in computation power of GPU and TPU, communication overhead becomes increasingly significant in DNN
training. It is especially important to minimize communication overhead when training using fragmented resources in
shared AI clusters, which are scattered on different machines or racks. Gradient compression has been used to reduce
tensor size for communication. Quantization \cite{quantization1} and sparsification \cite{sparsification1} are the most
used gradient compression methods \cite{dgc}, but introduce precision loss. We focus on SFB, which can reduce
communication time without affecting training convergence.

\subsection{Challenges}

\textbf{A very large strategy space.}
Suppose that we are deploying a DNN model of $n$ operations to a cluster of $m$ GPUs. We can replicate each op on any
subset of the $m$ devices, forming $2^m - 1$ choices. Also, for each replicated parameter (assuming there are $n_p$
replicated parameters), we choose between AllReduce or PS architecture for its parameter synchronization. In total
$(2^m-1)^n + 2^{n_p}$ strategies are to be explored.

\textbf{RL methods often suffer from overfitting.}
Reinforcement learning (RL) has shown success in exploring large search spaces \cite{confuciux,regal}. However, RL-based
methods tend to overfit the training data and may lead to poor generalization performance
\cite{rloverfit1,rloverfit2,go}. Conventional regularization methods such as data augmentation \cite{rloverfit1} hardly
help because the models and device topologies used in practice may differ drastically from those used in training. 

\textbf{SFB is not always beneficial.}
Communication cost of SFB depends on the number of replicas and size of sufficient factors (which is highly related to
the batch size). The communication-minimization choice among using SFB or a gradient synchornization method (AllReduce
or PS) needs to be examined for each gradient.

\subsection{Solutions}

To exploit the above opportunities and address the challenges, we propose \OurSystem, an automatic DNN deployment
framework, with the following key designs.

\textbf{Interactive strategy refinement with runtime feedback.}
Existing schemes like GDP \cite{gdp} and HeteroG \cite{heterog} generate strategies in a one-shot way: their models
directly output the strategy for the whole graph. Though the results are impressive, it is hard to interpret and reason
about the decision process. On the contrary, human developers often optimize distributed training strategies iteratively
and interactively. A developer may first run a simple strategy and examine the execution trace to find out the
bottleneck of this strategy and improve it accordingly. For example, if a strategy results in low utilization and long
idle time on one GPU, it may be worth considering putting more ops onto this GPU. Sec.5 in \cite{byteps} gives a
concrete example of such practice. Based on this observation, we design an automatic, interactive strategy-making
process in \OurSystem. Instead of taking only the raw features of a model as input and directly generating a strategy,
\OurSystem repeatedly simulates the execution of a strategy and collects the corresponding simulated execution trace,
and then tries to generate a better strategy based on the runtime feedback for the existing strategy. The interactive
strategy exploration process also helps \OurSystem produce feasible (e.g. not causing OOM errors) strategies for unseen
models. With systems that generate strategies in one-shot, if the generated strategy causes OOM, user intervention is
required to tune the parameters and retry. With interactive strategy exploration, \OurSystem automatically tries to
reduce the memory usage by more aggressive model parallelism when OOM errors are encountered, until a feasible solution
is found.

\textbf{Combining learning and searching.}
Learning-based methods require that datasets used for training should follow the same distribution as the real use
cases, or overfitting of the training data may occur \cite{rloverfit1,rloverfit2}. However, it is very hard to collect a
dataset that can represent the distribution of ``DNN models''. For example, PlaceTo \cite{placeto} is trained on only 3
models, GDP \cite{gdp} is learned on 11 models, and HeteroG \cite{heterog} is trained on 8 models. As new DNN models
emerge, a new model can differ drastically from those in the small training datasets. To mitigate this problem, we argue
that the learning-based methods should be paired with a search method that can robustly handle any unseen models. In
\OurSystem, we combine RL and MCTS, where the RL agent uses a GNN to produce prior probabilities to guide the MCTS
process.

\textbf{Unified representation of computation graph and device topology.}
Recent works have adopted GNN to produce distributed training strategies for different computation graphs
\cite{placeto,gdp,heterog}. Parameter dimensions of GNNs are not dependent on the number of nodes in a graph, allowing
them to generalize to graphs of different sizes. However, only the computation graph is encoded in the input to the GNN
in existing works, and their designs cannot adapt to unseen device topologies without retraining. To enable \OurSystem
to generalize to different device topologies, we use a heterogeneous graph as input, that contains both computation
nodes and device nodes. This allows us to encode all inputs in a unified graph and learn a heterogeneous GNN that can
generalize to both unseen DNN models and device topologies.

\textbf{Formulating SFB as an optimization problem.}
SFB deals with gradients of the parameters, whose only consumer is the \texttt{ApplyGradient} op. Whether or not to
apply SFB for a gradient is a local decision that is independent of the strategies of other ops. We formulate the SFB
decision for each gradient as a mixed-integer optimization problem that can be efficiently solved using off-the-shelf
solvers. The scale of the optimization problem remains small even for large graphs because it only focus on the subgraph
around a gradient.

\section{System Design}\label{sec:design}

\begin{figure}[!t]
    \centering
    \includegraphics[width=0.72\columnwidth]{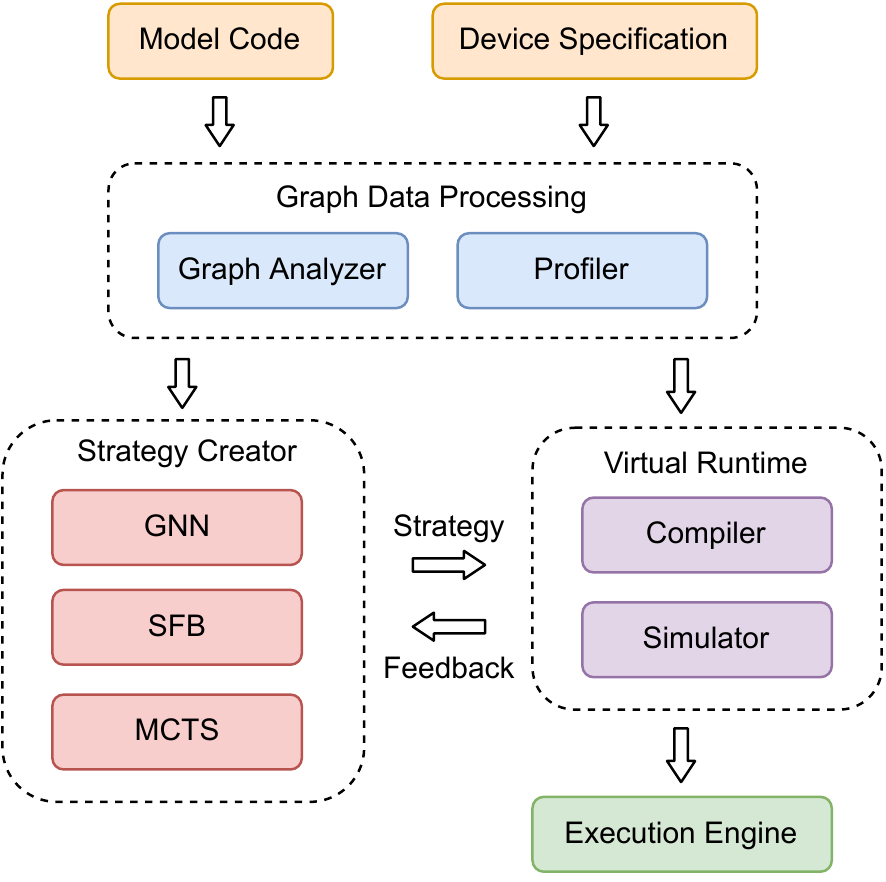}
    \caption{Workflow of \OurSystem.\label{fig:arch}}
    \vskip -1em
\end{figure}

The workflow of \OurSystem is given in Fig.~\ref{fig:arch}. A user first builds a single-GPU DNN computation graph using
a DL platform API, such as that provided in TensorFlow. The computation graph is fed to \OurSystem along with the device
specification, describing the available GPUs allocated for training this DNN. Though we focus on distributed training in
this paper, \OurSystem supports both training and inference tasks. For training jobs, the input computation graph should
contain both forward and backward ops (e.g., generated by TensorFlow's automatic differentiation engine). For inference
tasks, the input computation graph only contains the ops in the forward pass.

\OurSystem first extracts necessary features from the computation graph and device specification. A profiler
collects computation time of each op on each type of GPUs and the collective communication performance among the GPUs.
With these data, \OurSystem iteratively invokes the strategy creator and the virtual runtime to find good deployment
strategies. \OurSystem does not rely on any domain knowledge nor distinguish specific operators except for some special
ones such as \texttt{Placeholder}. It is hence generic and can support user-defined operators.

The strategy creator identifies a deployment strategy for a given DNN model. In the virtual runtime module, the compiler
modifies the computation graph accordingly, inserting necessary auxiliary ops that ensure the modified graph is
equivalent to the original graph. The simulator simulates the execution of the modified graph and estimates the training
time as feedback for the strategy creator to produce new and potentially better strategies.

\subsection{Graph Data Processing}

\subsubsection{Graph Analyzer}

The graph analyzer builds an internal representation of the original computation graph, that is independent of the API
used.

\vspace{1mm}
\textbf{Simplifying the graph.} Unnecessary nodes are removed including
\texttt{identity}, \texttt{NoOp} and the dangling ops that are not connected to the main
optimization ops (e.g., \texttt{GradientDescent}, \texttt{Adam}).
This reduces the graph size without changing the semantics.

\vspace{1mm}
\textbf{Annotating tensor split and concatenation methods.}
If two producer-consumer ops are replicated on different numbers of devices, split and concatenation ops are added:
replicated tensors are concatenated before being sent to an op that is not replicated, and a full tensor can be split
with different parts sent to different replicas of a replicated op.
However, only some ops can be split and concatenated while preserving mathematical equivalence.
The graph analyzer annotates each op with its splittability that will later be used by the compiler.
Specifically, the graph analyzer marks every op into one of the following categories. 

$\bullet$ {\em Splittable with concatenation.} Ops in this category can accept input tensors that are split in
the batch dimension. When the input is split, the output of such an op is also split and can be
concatenated in the batch dimension to recover the full tensor. Such ops
include element-wise ops (e.g., \texttt{AddN}), batched \texttt{Conv2D} and \texttt{MaxPool2D}, etc.

$\bullet$ {\em Splittable with element-wise summation.} Ops in this category can also accept split input tensors, but
the output needs to be summed, instead of being concatenated, to recover the full output tensor. This category typically
includes ops that produce gradients, e.g., \texttt{Conv2DBackpropFilter}. 

$\bullet$ {\em Others.} These ops do not accept split tensors. If the producer op is replicated, input tensors must
be aggregated to recover the full tensor, before being consumed by such an op. Specifically, \OurSystem marks
\texttt{ApplyGradient} ops into this category, so that gradients are automatically aggregated before being applied
to parameters.

\vspace{1mm}
\textbf{Grouping ops.} The number of ops varies a lot across DNN models. For example, the VGG model implemented in
TensorFlow Slim \cite{tfslim} has 1169 nodes, while a typical implementation of BERT-Large has 26601 nodes. To
efficiently produce strategies for models of different sizes, the graph analyzer employs METIS \cite{metis} to group
some tightly coupled ops together. We use METIS to partition the computation graph to no more than 60 groups by
minimizing the tensor sizes on the cut edges, while keeping the total computation time of each partition balanced with a
balance factor of 2. Since the replication or placement strategies of different op groups may differ, additional
communication is needed to aggregate and distribute the tensors on the op group boundaries. METIS minimizes the size of
these tensors, and hence minimizes this communication overhead. Each op group is regarded as a single node in the graph
passed to the strategy creator. A larger number of op groups allow more fine-grained strategies, at the cost of
lengthened searching time. We find that 60 groups achieve a good trade-off in our experiments.

\subsubsection{Profiler}

To measure computation time of each op, the profiler runs single-GPU model training on each type of GPUs. Since ops can
be replicated, we need the computation time under different batch sizes. We profile op execution time using typical
batch sizes below 60.
It is shown \cite{bao2020preemptive} that when the batch size is large enough,
computation time is almost linear with the batch size. We build a linear model to predict the computation time for
larger batch sizes that are not profiled. For a large model that does not fit in a single GPU even with small batch
sizes, we manually partition the model and profile each part separately. 

To predict the tensor transfer time, the profiler measures performance for GRPC transfer (between pairs of devices) and
NCCL AllReduce communication (among different combinations of devices). Random 32-bit floating-point-number data of
different sizes are transferred (starting from 1KB and doubled until 1GB). Segmented linear regression models are built
for GRPC transfer and for AllReduce communication.

\subsection{Strategy Creator}

The strategy creator consists of three components. {\em MCTS} progressively generates placement and replication
strategies for each op group. During the search, a heterogeneous {\em GNN} is exploited to provide prior probabilities
for MCTS to sample the strategy candidates, based on a graph input that joins the computation graph with the device
topology graph. When MCTS produces a strategy that replicates a parameter, {\em SFB solver} decides if SFB can be
applied to reduce communication.

Input to the strategy creator includes a computation graph and a device graph. The former contains $N$ nodes, each
representing an op group. The latter has $M$ nodes, each denoting a group of homogeneous GPUs, i.e., a set of GPUs of
the same type with the same bandwidth between each pair. This usually maps to a machine equipped with multiple same-type
GPUs.

The strategy creator produces a deployment strategy, containing op placements and the replication plan for each
replicated op. The placement $P$ is an $N \times M$ binary matrix. $P_{i,j} = 1$ if the i-th op group is placed on one
or all devices in the j-th device group (depending on the replication plan). The replication plan $O$ is a $N \times 4$
matrix where the i-th row corresponds to a one-hot encoding of 4 replication options for the i-th op group. $O_i$
defines how to place the i-th op group on the respective device group, if the device group includes multiple devices.
The 4 replication options considered in \OurSystem are:

$\bullet$ \textbf{Replicate with \texttt{AllReduce}.} The op group is replicated to all devices in the device group and
    input tensors are evenly split in the batch dimension. If the op group produces gradients, \texttt{AllReduce} ops
    are used for synchronization.
    
$\bullet$ \textbf{Replicate with PS.} It is similar to replicate with \texttt{AllReduce} except for using PS to
    synchronize gradients. The PS is chosen among GPUs in the device group in a round-robin manner.
    
$\bullet$ \textbf{Duplicate.} The op group will be copied to all devices in the device group. Unlike replication, input
    tensors are broadcast to all copies, so that gradients produced on all devices are identical, eliminating the need
    of synchronization.
    
$\bullet$ \textbf{Model Parallelism.} The ops in the op group are divided into smaller groups using METIS and placed to
    different devices in the device group. It is useful to place large models that otherwise cause out-of-memory (OOM)
    errors.

\subsubsection{Heterogeneous GNN}
\label{sec:heter_GNN}

\begin{figure*}[t]
    \vskip -.5em
    \centering
    \includegraphics[width=0.88\textwidth]{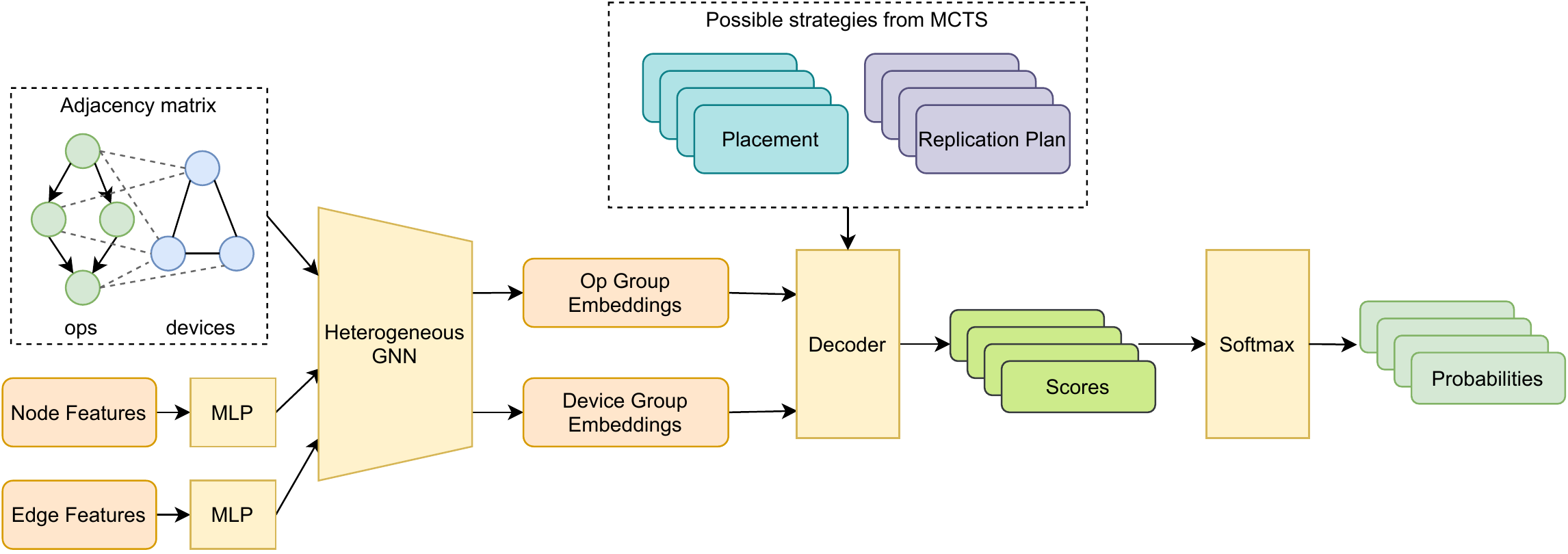}
    \vskip -1em
    \caption{Structure of the heterogeneous GNN.\label{fig:gnn}} 
    \vskip -1em
\end{figure*}

The GNN takes as input a heterogeneous graph, containing two types of nodes and three types of links. The first node
type is computation node, each representing a group of ops; the other type is device node, representing a homogeneous
group of GPUs. The three link types include: (i) links connecting two computation nodes, corresponding to tensors in the
computation graph; (ii) links that connect device nodes, representing a network link or a PCI switch; (iii) links
connecting computation nodes and device nodes, denoting a specific placement of the respective op group in the device
group.

Input features to the GNN contain four parts: (1) Raw features of the computation graph and devices, including total
computation time (averaged over measured running time on different devices) and overall parameter size of each op group,
the number of GPUs in each device group, memory capacity of each GPU, bandwidth between GPUs in each device group, size
of tensors connecting two op groups, and inter-group bandwidth between each pair of device groups. (2) A strategy
encoding, the one-hot encoding of replication plans of all op groups, and a binary edge feature for each edge that
connects an op group and a device group, indicating whether the former is placed on the latter. (3) Runtime feedback for
the input strategy, including makespan of each op group's execution, average idle time between the end of an op group's
execution and the start of its output tensor transfer, peak memory usage and idling percentage of each device group, and
idling percentage of each link between device groups. (4) The search progress, a one-hot encoding indicating which op
groups' deployment strategies have been decided and which op group's strategy will be produced next. We add
fully-connected layers to transform node-related features and edge-related features to feature vectors of fixed length
$f$ before feeding them to the GNN, so that the embedding lengths remain the same through GNN convolutions. We summarize
the input features in Table~\ref{tab:gnn_input}.

\begin{table}[t]
    \caption{GNN input features.\label{tab:gnn_input}}
    \vspace{-.5em}
    \centering
    \begin{tabular}{ | l | l | }
        \hline
        \textbf{Type}                & \textbf{Feature} \\ \hline
        \multirow{7}{*}{op node}     & computation time \\ \cline{2-2}
                                     & parameter size \\ \cline{2-2}
                                     & replication plan \\ \cline{2-2}
                                     & makespan \\ \cline{2-2}
                                     & idle time before transfering output \\ \cline{2-2}
                                     & if the strategy has been decided \\ \cline{2-2}
                                     & if the strategy is to be produced next \\ \hline
        \multirow{5}{*}{device node} & number of GPUs in the group \\ \cline{2-2}
                                     & memory capacity of each GPU \\ \cline{2-2}
                                     & intra-group bandwidth \\ \cline{2-2}
                                     & peak memory usage \\ \cline{2-2}
                                     & idling percentage \\ \hline
        op-op edge & tensor size \\ \hline
        \multirow{2}{*}{\makecell[l]{device-device \\ edge}} & inter-group bandwidth \\ \cline{2-2}
                                     & idling percentage \\ \hline
        op-device edge & placement \\ \hline
    \end{tabular}
    \vskip -1.5em
\end{table}

We adopt a 4-layer GNN. Each GNN layer transforms the embeddings produced by the previous layer by $ h^{i+1}_u =
\text{AGG}_{v \in \mathcal{N}(v)} \gamma_{etype} \cdot \sigma(W_{i,etype} (h^i_v \circ e_{uv}) + b_{i,etype})$. Here
$h^i_u$ is the embedding of node $u$ at the i-th layer, $\mathcal{N}(v)$ is the set of neighbors of node $v$ in the
input heterogeneous graph, and $e_{uv}$ is the edge feature between $u$ and $v$. $\circ$ is vector concatenation.
$W_{i,etype}$ and $b_{i,etype}$ are parameters of the i-th layer for $etype$ and $\sigma$ is a non-linear function.
$\text{AGG}$ represents the aggregation of features from neighbors. $h^0_u$ is initialized as the node features of node
$u$. $e$ It is observed that deep GNNs suffer from the over-smoothing issue \cite{oversmoothing1,oversmoothing2}, i.e.,
deeper GNNs do not necessarily perform better. Our experience confirms this phenomenon and we find that 4 layers give
the best results in \OurSystem. We choose multi-head attention based aggregation introduced in graph attention networks
(GAT) \cite{gat} as it assigns different weights to different neighbors, which is desirable because we believe that some
of the neighbors are more important in determining the best strategy for an op group. $\gamma_{etype}$ is the weight for
different types of edges: $\gamma_{etype}$ is set to $1$ for edges connecting the same types of nodes and $0.1$ for
those connecting different types of nodes, so as to balance the different types of edges connecting a node.
Fig.~\ref{fig:gnn} shows the structure of the GNN.

Output of the GNN includes embeddings $E_{op}$ of op groups and embeddings $E_{dev}$ of device groups. The GNN is
further connected to a thin decoder, which contains a simple \texttt{Dense} layer that takes as input a strategy slice
($P_i$, $O_i$), containing placement and replication plan of the i-th op group, and feature vector $\sum_{j=1}^M
E_{dev}[j]P_{i,j} \circ E_{op}[i] \circ O_i $ (involving embeddings produced by the GNN). It computes a score for the
strategy slice. A softmax op further produces probabilities for all possible strategy slices of the i-th op group
according to their scores. The probabilities are used for guiding MCTS's exploration.

\subsubsection{Monte-Carlo Tree Search}

MCTS is a best-first search algorithm that balances exploitation and exploration. In our search, a vertex in the search
tree represents a partial deployment strategy $s$ and an edge is an action $a$. The partial strategy includes incomplete
placement and replication plan matrices $P$ and $O$, i.e., the matrices with some rows filled, corresponding to some op
groups. $a$ is the deployment strategy to be applied to the next op group in consideration. Op groups are sorted in
descending order of computation time, so that the most computation-expensive op group will be considered first. Each
edge records its visit count $N(s, a)$ and a running average reward $Q(s, a)$. We use the simulator to estimate
execution time of the DNN graph with the current deployment strategy and calculate the speed-up over the baseline
strategy (aka DP with AllReduce-based parameter synchronization) as the reward.

MCTS starts with an empty strategy $s_0$, and progressively builds a search tree by repeatedly performing the following:

$\bullet$ {\em Selection}: Starting from the root of the tree, keep selecting child vertices to traverse, until we reach
a leaf vertex (a vertex which has not been expanded, or with complete deployment strategies for all op groups). At each
vertex $s$, the edge $(s,a)$ with the highest PUCT score \cite{puct} is picked to traverse next, as defined as: 

\vspace{-1em}
$$ U(s, a) = Q(s, a) + cG(s, a)\frac{\sqrt{\sum_{a^{\prime} \in \phi} N(s, a^{\prime})}}{1 + N(s, a)} $$
\vspace{-1em}
    
\noindent where $c$ is a coefficient, $G(s, a)$ is the prior probability produced by the GNN, and $\phi$ denotes all
actions in the child vertices of $s$. The PUCT score prioritizes the most promising strategies for exploration. 

$\bullet$ {\em Expansion and Evaluation}: When reaching a leaf vertex, we evaluate the reward $r$ of this vertex, which
is the training speed-up achieved by its strategy over the baseline (DP), using the simulator.\footnote{For op groups
whose deployment strategies have not been decided, we use the strategy of the most computation-expensive op group on
them.} If an OOM error results, the reward is set to $-1$. Then we expand the subtree by one level, by enumerating
possible strategies for the next op group and obtain their prior probabilities $G(s, a)$ from the GNN.

$\bullet$ {\em Back-propagation}: After obtaining reward $r$ at a leaf vertex, the running average reward $Q(s,a)$ and
the visit count $N(s,a)$ along the path from the root vertex to the leaf vertex are updated, e.g., $Q(s,a) := Q(s,a) +
\frac{1}{N(s,a)} \cdot r$. 
\vspace{1mm}

\begin{figure}[t]
    \centering
    \includegraphics[width=0.85\columnwidth]{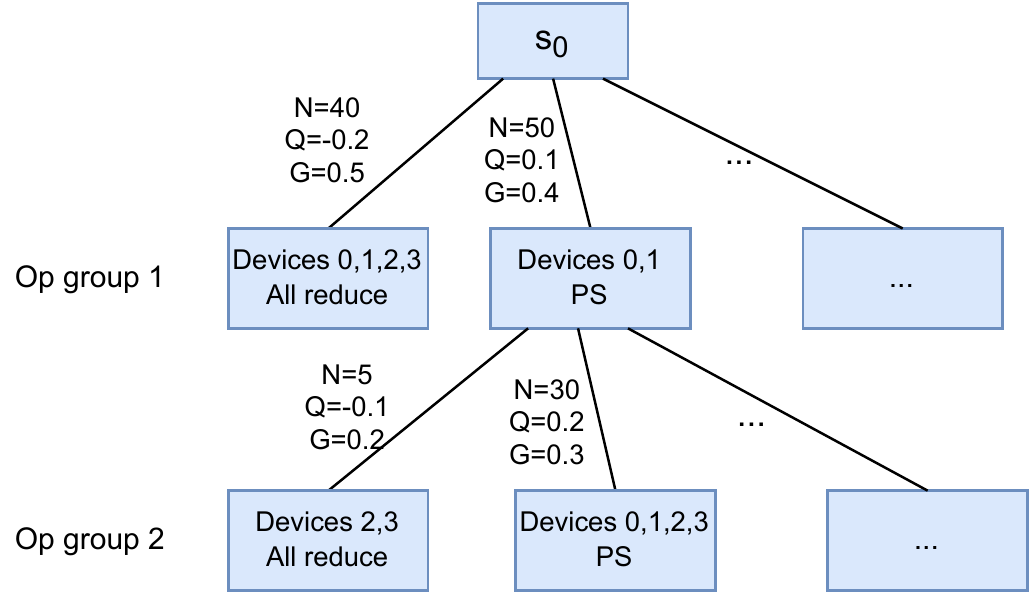}
    \caption{Example of an MCTS search tree.\label{fig:mcts}}
    \vspace{-1em}
\end{figure}

Fig.~\ref{fig:mcts} shows an example of the MCTS search tree. The root node is the empty strategy
$s_0$. Each level includes the placement and replication plan for an op group. In each iteration, the search tree is
traversed from the root node to a leaf node according to the selection policy. Then it expands one level and evaluates
the reward to update the $Q$ and $N$ on the path.

We choose MCTS over other searching methods because of the following: {\em First}, it allows natural integration with a
RL-based method (for updating GNN model parameters), which is important for generalizability. In each step of GNN
training, we randomly choose a DNN graph and a device topology. We run MCTS and collect the selection probability
$\pi(s) = \operatorname{softmax}\ln N(s)$ at vertices with at least 800 visit counts, where $N(s)$ is a vector including
visit counts of all child vertices of $s$. Parameters $\theta$ are then updated to minimize the cross entropy between
the prior probability $G_\theta(s, a)$ produced by the GNN and the selection probability $\pi(s, a)$ of the MCTS. {\em
Second}, MCTS builds the strategy progressively. At each vertex, we evaluate the partial strategy and use the runtime
feedback to help GNN make better prediction for further strategies.

\subsubsection{Sufficient Factor Broadcasting}

When a parameter is replicated on multiple devices (with the corresponding op), its gradients need to be synchronized
across the devices. This is usually achieved by inserting an \texttt{AllReduce} or PS op. Alternatively, some gradients
can be calculated from tensors of small sizes. For example, a non-full-rank gradient matrix can be represented by the
product of two smaller matrices (the small matrices are the sufficient factors). Our SFB solver is designed to
automatically find these mathematically equivalent replacements in the graph that can potentially reduce communication.

\begin{figure}[t]
    \centering
    \subfloat[Example cuts for SFB.]{%
        \includegraphics[width=0.65\columnwidth]{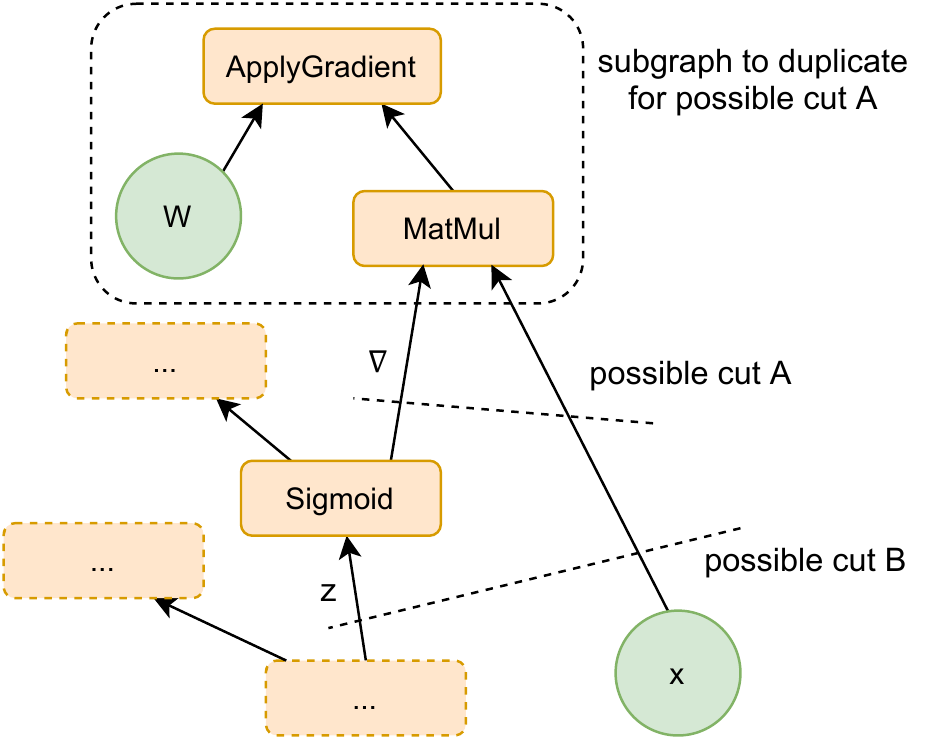}%
    }
    \vskip .05in
    \subfloat[Implementation of SFB for cut A in (a).]{%
        \includegraphics[width=\columnwidth]{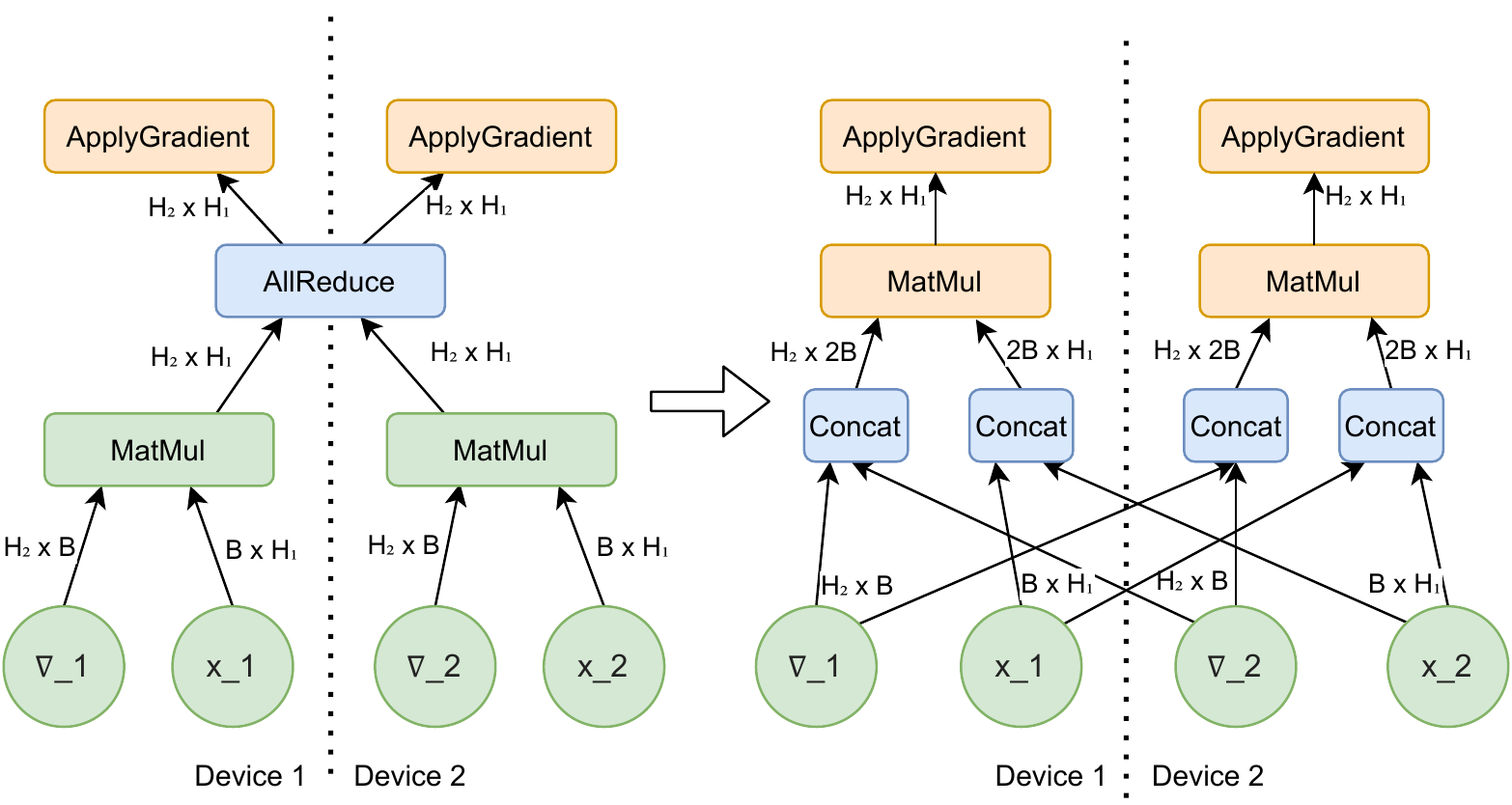}%
    }
    \vspace{-0.2em}
    \caption{An example of sufficient factor broadcasting.\label{fig:sfb_impl}}
    \vspace{-1em}
\end{figure}

Fig.~\ref{fig:sfb_impl}(b) shows an example of applying SFB to an \texttt{MatMul} op that can be found in the
\texttt{Dense} layers in many DNNs. When \texttt{MatMul} is replicated onto multiple devices, instead of using
\texttt{AllReduce} or PS to synchronize produced gradients, the sufficient factors, $\nabla$ and $x$, are broadcast to
all devices, and \texttt{MatMul} ops on each device can reconstruct identical gradients. In \OurSystem, this corresponds
to choosing the ``Duplicate'' option for the respective op as the deployment strategy. It changes the total
communication data from the gradient size, $H_2 \times H_1$, to the size of the sufficient factors, $2(H_2 \times B + B
\times H_1)$, where $H_1$ and $H_2$ are the input and output feature lengths of the \texttt{Dense} layer and $B$ is the
batch size. Our SFB optimization identifies gradients where this change is beneficial.

A DNN computation graph includes optimizer ops that update parameters in the model. Consider such an op $l$ (e.g.,
\texttt{ApplyGradient} in Fig.~\ref{fig:sfb_impl}). One of its input op $g$ (e.g., \texttt{MatMul} in
Fig.~\ref{fig:sfb_impl}) produces the gradient of the parameter that $l$ updates. If the op group containing $g$ is
replicated by MCTS and tensor $(g,l)$ needs to be synchronized among all replica devices, we check if SFB can be applied
to reduce communication. Note that MCTS can also directly produce ``Duplicate'' option to enable SFB on some op group.
Here we are double checking opportunities of applying SFB on op groups for which MCTS has produced a replication option.
The rationale lies in that MCTS produces strategies on the op group level and the group boundaries decided by METIS are
rarely the best cuts for SFB.

For every gradient tensor in a replicated op group, we solve the following optimization problem to determine a subgraph
(e.g., the dashed box in Fig.~\ref{fig:sfb_impl}(a)) that can be duplicated. Since sufficient factors contain all inputs
used to calculate a respective gradient, they form a cut that separates the subgraph from the rest of the computation
graph. For example, in Fig.~\ref{fig:sfb_impl}(a), $\nabla$ and $x$ are a set of sufficient factors of the gradient
produced by \texttt{MatMul} and they form a cut; the gradient of $W$ can be calculated by running the subgraph based
only on $\nabla$ and $x$. Existing studies \cite{sfb,zhang2017poseidon} only consider SFB for ops that matmul two
vectors, i.e., gradients that are outer products of two vectors; our approach can identify other cases as long as such
cuts can be found.

\noindent\begin{minipage}{\columnwidth}
\small
\begin{equation}
    \min~~ (D-1) \sum_{i \in V}
    \alpha_i T_i + D(D-1)\hspace{-2mm}\sum_{(j,i) \in E}\hspace{-1mm}b_{ji}\frac{L_{ji}}{\tau}
                 - 2\alpha_g \frac{D\hspace{-.8mm}-\hspace{-.8mm}1}{D} \frac{L_{gl}}{\tau} \nonumber
\end{equation}
\vspace{-3mm}
\begin{align*}
    \text{subject to:} &\quad \alpha_k \leq \sum_{(k,i) \in E}\alpha_i, \quad\forall k \in V \setminus \{l\} \\[-1mm]
                       &\quad b_{ji} \geq \alpha_i - \alpha_j, \quad\forall (j,i) \in E \\
                       &\quad \alpha_i \in \{0, 1\}, \quad\forall i \in V \\
                       &\quad b_{ji} \in \{0, 1\}, \quad\forall (j,i) \in E
\end{align*}
\end{minipage}
\vskip 1em

\begin{table}[t]
    \caption{\label{tab:sfb_symbols}Notation in SFB Optimization.}
    \vspace{-.5em}
    \begin{center}
    \begin{small}
    \scalebox{0.95}{
    \begin{tabular}{ | c | l | } \hline
        \rule{0pt}{1em}$E$ & the set of tensors inside the op group \\[.1em] \hline
        \rule{0pt}{1em}$V$ & the set of ops in the op group \\[.1em] \hline
        \rule{0pt}{1em}$l$ & an optimizer op \\[.1em] \hline
        \rule{0pt}{1em}$g$ & an op that produces gradient for $l$ \\[.1em] \hline
        \rule{0pt}{1em}$D$ & the number of devices that have a replica of $g$ \\[.1em]  \hline
        \rule{0pt}{1em}$T_i$ & computation time of op $i$ \\[.1em] \hline
        \rule{0pt}{1em}$L_{ji}$ & size of tensor $(j,i)$ \\[.1em] \hline
        \rule{0pt}{1em}$\tau$ & bottleneck bandwidth between the $D$ devices \\[.1em]  \hline
        \rule{0pt}{1em}$\alpha_i$ & whether or not to duplicate op $i$ \\[.1em] \hline
        \rule{0pt}{1em}$b_{ji}$ & whether tensor $(j,i)$ is in the cut or not \\[.1em] \hline
    \end{tabular}
    }
    \end{small}
    \end{center}
    \vspace{-1.5em}
\end{table}

Notation is given in Table~\ref{tab:sfb_symbols}. $\alpha_i$'s are the main decisions: $\alpha_i = 1$ means changing the
replication option of op $i$ from ``Replicate with AllReduce'' or ``Replicate with PS'' to ``Duplicate''; and $\alpha_i
= 0$, otherwise. $b_{ji}$ specifies if tensor (j,i) is in the cut that partitions the subgraph containing op $i$ out.

The first term in the objective is the extra computation time due to duplication. Since we duplicate the ops in all
devices where the ops' replicas are placed, each device needs to process the ops $D-1$ more times than not to duplicate.
The second term minus the third term is the extra communication time incurred. For every tensor that is produced by a
replicated op and consumed by a duplicated op, $D(D-1)$ transfers of the tensor are needed to broadcast it to all
devices involved in the duplication. If no duplication is applied, the gradient can be synchronized with an AllReduce or
PS method, and the third term formulates the communication time of using ring AllReduce (as an example, and the case of
other AllReduce algorithms and PS can be formulated accordingly). The first constraint specifies that an op is included
in the duplicated subgraph only if one of its consumer ops is in it, as otherwise it is unrelated to the gradient. The
second constraint ensures that a tensor $(j,i)$ is in the cut if $i$ is duplicated and $j$ is not. It is an integer
linear program similar to the min-cut problem, but with additional node weights on one side of the cut. We use the Cbc
\cite{cbc} solver to solve the problem.

\subsection{Virtual Runtime}

The virtual runtime evaluates a strategy without actually running it on a physical cluster. It also generates the
distributed training graph according to the strategy, which can then be loaded and executed by the execution engine
(e.g. TensorFlow).

\subsubsection{Compiler}

The compiler takes as input the DNN computation graph and a deployment strategy $(P, O)$ found by the strategy creator.
It applies the strategy and produces a modified computation graph for either the simulator or the actual execution
engine, e.g., TensorFlow. The compiler automatically inserts necessary auxiliary ops to ensure the equivalence of the
modified graph and the original graph regardless of the deployment strategy. It first maps the ops to devices according
to the placement $P$, and then inserts auxiliary ops in the following cases: 

$\bullet$ When an op is replicated but its input tensors are not, \texttt{Split} ops are inserted to split the input
tensors, before feeding them to the op's replicas.

$\bullet$ When an op is not replicated but its input tensors are replicated, \texttt{Concat} or \texttt{AddN} ops are
added to aggregate input tensors, used for parent ops marked as ``Splittable with concatenation'' and ``Splittable with
element-wise summation'', respectively.

$\bullet$ When both the op and input tensors are replicated but on different numbers of devices, both \texttt{Concat}
and \texttt{Split} ops are inserted to adjust the number of replicas.

$\bullet$ When a parameter is replicated, \texttt{AllReduce} or \texttt{AddN} op is inserted depending on the
replication option.

\subsubsection{Simulator}

The simulator implements an op scheduling algorithm similar to the default scheduler of TensorFlow. It sets up a FIFO
queue for each device. When all input tensors of an op is ready, the op is inserted to the queue. Each device
independently simulates the execution of the ops in its task queue using the profiled data and reports the finish time
of each op.

The simulator uses reference counting to track the lifetime of tensors and estimate the peak memory usage on each
device. When an op is done, its output tensors are added into memory usage of the respective device. Once all
ops that use a tensor are executed, the tensor is considered de-allocated and removed from the device memory
usage.
\section{Implementation and Evaluation}\label{sec:evaluation}

\subsection{Implementation}

We implement \OurSystem as a Python module on TensorFlow 1.14. For op grouping, we use tensor size as the edge weight
and computation time as node balancing constraints when running METIS. We set a default partition number of 60 in our
experiments. The profiler runs TensorFlow with tracing options to collect computation time of each op under different
batch sizes. With each batch size, each model is profiled 5 times to obtain the average time.

Our heterogeneous GNN is implemented on DGL \cite{dgl}. We extend the GAT \cite{gat} implementation in DGL to support
heterogeneous graphs and edge features. The GNN has 55MB of parameters. We use Cbc \cite{cbc} to solve the SFB
optimization problem. In our experiments, we find that it can reliably solve the integer optimization problem within
hundreds of milliseconds.

\subsection{Experimental Set-up\label{sec:exp_setup}}

\textbf{Hardware.} We conduct the experiments on two clusters. The first cluster is an on-premise cluster
(\textit{testbed}) of 7 physical machines: one is equipped with 4 NVIDIA Tesla V100 32GB GPUs; four are each equipped
with 2 NVIDIA GTX 1080Ti GPUs; the other two are each equipped with 2 NVIDIA Tesla P100 GPUs. The first machine has
NVLink, while PCIe is used on the other machines. The machines are connected to a 100Gbps switch. The second cluster is
on a public cloud (\textit{cloud}) with 6 machines and 32 GPUs. Two of the machines are equipped with 8 NVIDIA Tesla
V100 16GB GPUs and the other 4 with 4 NVIDIA Tesla T4 GPUs. These machines are inter-connected with 10Gbps bandwidth.

\vspace{1mm}
\textbf{Benchmarks.} We experiment with 6 representative DNN models for image classification and neural language
processing, as listed in Table~\ref{tab:models}. Adam\cite{adam} optimizer is used in the experiments.

\begin{table}[t]
    \caption{\label{tab:models}Benchmark DNN models.}
    \begin{center}
    \begin{small}
    \vskip -1em
    \scalebox{0.95}{
    \begin{tabular}{|l|r|r|r|}\hline
        \textbf{Model} & \textbf{Batch size} & \textbf{\# of ops} & \textbf{Parameter size} \\\hline
        InceptionV3\cite{inception}   &  96 & 5312  & 90M \\\hline
        ResNet101\cite{resnet}        &  96 & 7951  & 169M \\\hline
        VGG19\cite{vgg}               &  96 & 1169  & 548M \\\hline
        Transformer\cite{transformer} & 480 & 16859 & 407M \\\hline
        BERT-Small\cite{bert}         &  96 & 5061  & 98M \\\hline
        BERT-Large\cite{bert}         &  16 & 26601 & 2313M \\\hline
    \end{tabular}
    }
    \end{small}
    \end{center}
    \vspace{-1.5em}
\end{table}

\vspace{1mm}
\textbf{GNN Training.} We train the GNN in strategy creator using the 6 DNN models, the testbed device topology and
randomly generated 100 device topologies as input. A random device topology is produced with a machine number in [1, 6],
[1, 8] GPUs per machine of a GPU type among 3 types, intra-machine bandwidth between [64, 160] Gbps (to simulate the
absence or presence of NVLink) and inter-machine bandwidth within [20, 50] Gbps (to reflect different machine
locations). It takes around 2 days for GNN training to converge.

\vspace{1mm}
\textbf{Baselines.} We compare \OurSystem with the following baselines.
(1) \textbf{DP-NCCL}: data parallelism with NCCL \cite{nccl} AllReduce for parameter synchronization.
It is widely used for distributed training and is built-in in most DL frameworks \cite{tensorflow,pytorch,mxnet}.
We implement DP-NCCL using standard in-graph replication on TensorFlow.
(2) \textbf{DP-NCCL-P}: data parallelism with batch sizes allocated to the devices being inverse propotional to their
computation capacities.
(3) \textbf{Horovod} \cite{horovod}: a data-parallel training framework (used with TensorFlow in our experiments)
that incorporates optimizations such as overlapping of AllReduce and backward computation.
(4) \textbf{FlexFlow} \cite{flexflow}: a distributed
deep learning framework that supports parallelization in the SOAP dimensions. FlexFlow assumes that the cluster is
homogeneous. We implemented VGG19, ResNet101 and InceptionV3 on it using its Keras API. Due to the lack of attention layer implementation,
we were not able to implement all our benchmark models on FlexFlow. We set the number of its MCMC search iterations to 100000.
(5) \textbf{HDP} \cite{hdp}: a heterogeneity-aware hierarchical device placement system that joinly learns grouping and device allocations with reinforcement learning.
(6) \textbf{Post} \cite{post}: a device placement system with cross-entropy minimization and proximal policy optimization.
(7) \textbf{PlaceTo} \cite{placeto}: a device placement system using GNN and reinforcement learning.
(8) \textbf{GDP} \cite{gdp}: a device placement system using GNN and Transformer models.
(9) \textbf{Baechi} \cite{baechi}: an algorithmic device placement system. We only compare with its \textit{mSCT} algorithm as it is reported to outperform the other two algorithms proposed in the paper.
(10) \textbf{HeteroG} \cite{heterog}: a state-of-the-art system that supports heterogeneous clusters, which uses a GNN to
make op deployment decisions. It supports a similar decision space as \OurSystem, but only assumes one given device
topology and replicates an op to all devices or put it on a single device. We train HeteroG with all the
benchmark models under the device topology of our testbed. 

Some of the baseline systems are not open-source and non-trivial to reimplement. We adopt the evaluation methodology
used in related work \cite{baechi,heterog} and compare the reported improvements over expert strategies for these
baselines.

\subsection{Training Speed-up on Heterogeneous Clusters}

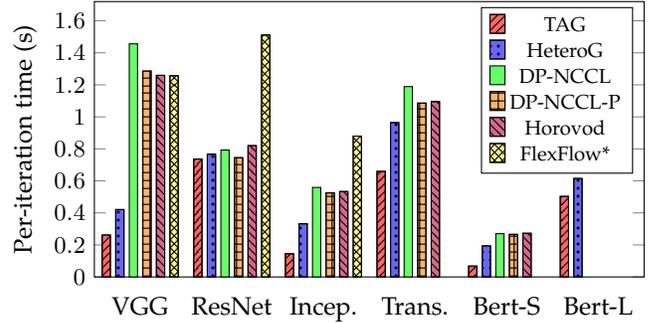
\begin{figure}[t]
    \vspace{-.5em}
    \centering
    \begin{tikzpicture}[scale=1, every node/.style={transform shape}]
        \begin{axis} [
            ybar,
            ymin=0,
            ymax=1.72,
            ytick distance=0.2,
            bar width=.11cm,
            width=1\columnwidth,
            height=.59\columnwidth,
            symbolic x coords={VGG, ResNet, Incep., Trans., Bert-S, Bert-L,},
            enlarge x limits=0.1,
            xtick style={draw=none},
            ylabel={Per-iteration time (s)},
            legend image code/.code={
                \draw [#1] (0cm,-0.1cm) rectangle (0.2cm,0.15cm);
            },
            legend style={nodes={scale=0.8, transform shape}}
        ]
            \addplot[style={fill=red!60,postaction={pattern=north east lines}}] coordinates { (VGG, 0.2618446944002062) (ResNet, 0.7365205493569374) (Incep., 0.1448051673406735) (Trans., 0.6594209679588675) (Bert-S, 0.0678673808183521) (Bert-L, 0.5032803978165612) };
            \addplot[style={fill=blue!60,postaction={pattern=dots}}] coordinates { (VGG, 0.421) (ResNet, 0.767) (Incep., 0.332) (Trans., 0.964) (Bert-S, 0.194) (Bert-L, 0.615) };
            \addplot[style={fill=green!60}] coordinates { (VGG, 1.4561474920995534) (ResNet, 0.7928568560397252) (Incep., 0.55929187042173) (Trans., 1.1889082144387066) (Bert-S, 0.2703741769399494) };
            \addplot[style={fill=orange!60},postaction={pattern=grid}] coordinates { (VGG, 1.286) (ResNet, 0.745) (Incep., 0.525) (Trans., 1.086) (Bert-S, 0.266) };
            \addplot[style={fill=purple!60,postaction={pattern=north west lines}}] coordinates { (VGG, 1.25925) (ResNet, 0.82002) (Incep., 0.53356) (Trans., 1.09493) (Bert-S, 0.27281) };
            \addplot[style={fill=yellow!60,postaction={pattern=crosshatch}}] coordinates { (VGG, 1.2573) (ResNet, 1.5112) (Incep., 0.8797) };
            \legend{\OurSystem, HeteroG, DP-NCCL, DP-NCCL-P, Horovod, FlexFlow*}
        \end{axis}

    \end{tikzpicture}
    \vspace{-.5em}
    \caption{Per-iteration training time. DP-NCCL, DP-NCCL-P, and Horovod result in OOM with BERT-Large. *FlexFlow is implemented on Legion \cite{legion} while other systems are implemented on TensorFlow \cite{tensorflow}.
    \label{fig:performance_comparison}}
\end{figure}

We first compare the average per-iteration training time incurred by \OurSystem and the open-source baselines on our
testbed (Sec.~\ref{sec:exp_setup}). Fig.~\ref{fig:performance_comparison} shows that \OurSystem outperforms the
baselines across all models: 8\%-456\% speed-up compared with DP-NCCL, 1\%-391\% speed-up compared with DP-NCCL-P,
11\%-381\% speed-up compared with Horovod, and 4\%-186\% speed-up compared to HeteroG. Note that HeteroG is trained from
scratch for this device topology. The best acceleration is achieved over DP-NCCL when training VGG19: VGG19 has a
relatively large number of parameters and communication tends to be the bottleneck; with DP-NCCL, parameter
synchronization happens after the slowest devices are ready, which slows down training substantially. Models such as
ResNet101, on the other hand, have less parameters but are more computation-intensive; DP-NCCL utilizes all GPUs well
and both \OurSystem and HeteroG can hardly find much better strategies. Horovod outperforms DP-NCCL in most cases due to
its highly optimized communication, but the speed-up is very limited in a heterogeneous cluster. DP-NCCL-P balances the
computation workload of different cards and performs better than DP-NCCL in the heterogeneous cluster. However, the
overall improvement is very limited due to two reasons. First, it has the same communication time as DP-NCCL: both
methods synchronize all parameters among all devices with AllReduce. Second, different operators have different
computation characteristics. The optimal batch size distribution among the devices varies for different operators.
FlexFlow finds hybrid strategies that choose different replication numbers and placements for different layers. However,
since FlexFlow does not consider the heterogeneity among computation devices, it puts an excessive amount of workload on
slow cards and results in suboptimal performance.

\begin{table}[t]
    \caption{\label{tab:strategy_stat} Deployment strategies produced by \OurSystem.}
    \vspace{-.5em}
    \begin{center}
    \begin{small}
    \scalebox{0.95}{
    \begin{tabular}{|l|r|r|r|r|r|}\hline
        \multirow{2}{*}{\textbf{Model}} &   \multicolumn{3}{c|}{\textbf{Replication}} & \multicolumn{2}{c|}{\textbf{Communication}} \\\cline{2-6}
                     &  \textbf{\textit{V100}} & \textbf{\textit{1080Ti}} & \textbf{\textit{P100}} &     \textbf{\textit{PS}} & \textbf{\textit{AR}}\\\hline
        InceptionV3  &   4.0 &    8.0 &  0.0 &  100\% &    0\% \\\hline
        VGG19        &   4.0 &    2.1 &  0.0 &  100\% &    0\% \\\hline
        ResNet101    &   4.0 &    8.0 &  4.0 &   67\% &   23\% \\\hline
        Transformer  &   4.0 &    8.0 &  0.0 &  1.7\% & 98.3\% \\\hline
        Bert-Small   &   3.9 &    0.1 &  0.1 & 15.0\% & 85.0\% \\\hline
        Bert-Large   &   0.6 &    0.5 &  0.1 &  3.5\% &    0\% \\\hline
    \end{tabular}
    }
    \end{small}
    \end{center}
    \vskip -2em
\end{table}

\begin{table*}[!t]
    \vspace{-.5em}
    \caption{Per-iteration training time (seconds) with and without applying sufficient factor broadcasting. \label{tab:sfb_exp}}
    \vspace{-.5em}
    \begin{center}
    \begin{small}
    \scalebox{0.95}{
    \begin{tabular}{|l|r|r|r|r|r|r|r|}\hline
        \multirow{2}{*}{\textbf{Model}} &      \multicolumn{3}{c|}{\textbf{DP-NCCL}}     &  \multicolumn{3}{c|}{\textbf{\OurSystem}} & \multirow{2}{*}{\textbf{FlexFlow}} \\\cline{2-7}
                               & \textbf{\textit{without SFB}} & \textbf{\textit{with SFB}} & \textbf{\textit{Speedup}} & \textbf{\textit{without SFB}} & \textbf{\textit{with SFB}} & \textbf{\textit{Speedup}} & \\\hline
        InceptionV3            & 0.0898      &   0.0452 &  98.7\% &      0.0775 &   0.0420 &  84.5\% & 0.0631 \\\hline
        ResNet101              & 0.1122      &   0.0854 &  31.4\% &      0.0508 &   0.0490 &   3.8\% & 0.0552 \\\hline
        VGG19                  & 0.1195      &   0.1192 &   0.3\% &      0.1167 &   0.1169 &  -0.2\% & 0.1083 \\\hline
        Transformer            & 0.1199      &   0.0455 & 163.5\% &      0.0521 &   0.0451 &  15.5\% & N/A \\\hline
        BERT-Small             & 0.0618      &   0.0546 &  13.2\% &      0.0597 &   0.0535 &  11.6\% & N/A \\\hline
        BERT-Large             & 0.4576      &   0.4317 &   6.0\% &      0.4331 &   0.4266 &   1.5\% & N/A \\\hline
    \end{tabular}
    }
    \end{small}
    \end{center}
    \vskip -1.5em
\end{table*}

Table~\ref{tab:strategy_stat} provides details of the strategies produced by \OurSystem, on the average number of GPUs
of each type that ops are replicated onto and percentages of gradients that use PS or AllReduce for synchronization. For
ResNet101, \OurSystem replicates all ops onto all devices; for other models, the 4 P100 GPUs are rarely exploited,
because benefits provided by using these devices do not cover the additional communication costs. We observe that a
mixture of PS and AllReduce is used for parameter synchronization when training most models; the ``duplicate'' option is
not selected because it is mainly effective with small batch sizes (as in the case to be presented in
Sec.~\ref{sec:SFB_exp}), and batch sizes used in this experiment are relatively large.

\subsection{Training Speed-up on Homogeneous Clusters\label{sec:eval_homogeneous}}

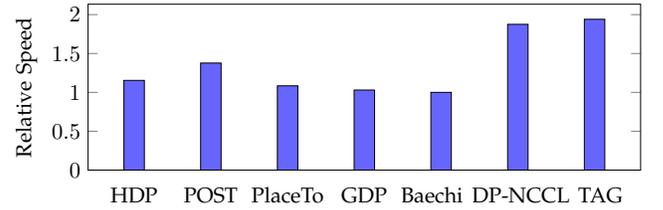
\begin{figure}[t]
    \vspace{-.5em}
    \centering
    \begin{tikzpicture}[scale=.85, every node/.style={transform shape}]
        \begin{axis} [
            ybar,
            ymin=0,
            bar width=.32cm,
            width=1.15\columnwidth,
            height=.47\columnwidth,
            symbolic x coords={HDP, POST, PlaceTo, GDP, Baechi~~~~, ~DP-NCCL, ~~\OurSystem, },
            xtick style={draw=none},
            tick label style={font=\fontsize{10pt}{1em}},
            ylabel={Relative Speed},
            xlabel=\phantom{Iter},
            xtick=data,
        ]
            \addplot[style={fill=blue!60}] coordinates {
                (HDP, 1.154)
                (POST, 1.377)
                (PlaceTo, 1.084)
                (GDP, 1.03)
                (Baechi~~~~, 1)
                (~DP-NCCL, 1.875)
                (~~\OurSystem, 1.94)
            };
        \end{axis}

    \end{tikzpicture}
    \vspace{-1.5em}
    \caption{Training speed of InceptionV3 on homogeneous clusters. The speed is relative to the human expert strategy.\label{fig:homogeneous}}
    \vspace{-1em}
\end{figure}

We compare \OurSystem with more baselines, including those not providing open-source implementation, by comparing
the reported speed-up over human expert strategies following the evaluation methodology in the related work
\cite{baechi,heterog}. For fair comparison, we conduct this experiment under a similar hardware setting as in these
works, which is a homogeneous cluster with two V100 GPUs on the same machine. We use InceptionV3 as the benchmark model
because it is the only common model evaluated in all these works. The same expert strategy as in existing studies
\cite{hdp,post} is used for the benchmark model. As shown in Fig.~\ref{fig:homogeneous}, \OurSystem outperforms 
all baselines by 3\%-94\%.

\subsection{Effectiveness of the Runtime Feedback Features\label{sec:eval_runtime_feedback}}

Among the four parts of feature input to our GNN (Sec.~\ref{sec:heter_GNN}), part 3 includes a number of features
provided by the simulator. Other studies which use a simulator to drive GNN learning (e.g., FlexFlow \cite{flexflow} and
HeteroG \cite{heterog}) largely exploit only execution time produced by the simulator, while we use multi-dimensional
information estimated by the simulator as GNN input. To evaluate effect of such features, we train the GNN with and
without them. Fig.~\ref{fig:feedback_learning_curve} shows that the runtime feedback features significantly boost the
learning of the GNN.

\begin{figure}[t]
    \begin{minipage}[t]{0.49\linewidth}
    \centering
    \begin{tikzpicture}[scale=0.88, every node/.style={transform shape}]
        \begin{axis}[
            xlabel=Iteration,
            ylabel=Cross Entropy,
            xmin=-1000,
            xmax=115000,
            xtick distance=20000,
            ytick distance=0.2,
            scaled x ticks=manual:{$\times 10^{4}$}{
                \pgfmathparse{#1/10000}
            },
            width=1.2\linewidth,
            height=\linewidth,
            legend image code/.code={
                \draw [#1] (0cm,0.08em) to (1em,0.08em);
            },
            legend style={nodes={scale=0.85, transform shape}}
            ]

        \addplot[color=red] coordinates { (0,2.843) (500,2.774) (1000,2.738) (1500,2.704) (2000,2.681) (2500,2.651) (3000,2.631) (3500,2.585) (4000,2.562) (4500,2.549) (5000,2.541) (5500,2.526) (6000,2.499) (6500,2.488) (7000,2.483) (7500,2.494) (8000,2.505) (8500,2.485) (9000,2.472) (9500,2.466) (10000,2.463) (10500,2.454) (11000,2.451) (11500,2.468) (12000,2.449) (12500,2.463) (13000,2.433) (13500,2.431) (14000,2.428) (14500,2.433) (15000,2.444) (15500,2.42) (16000,2.416) (16500,2.429) (17000,2.413) (17500,2.421) (18000,2.385) (18500,2.373) (19000,2.375) (19500,2.394) (20000,2.412) (20500,2.407) (21000,2.414) (21500,2.42) (22000,2.421) (22500,2.411) (23000,2.426) (23500,2.443) (24000,2.463) (24500,2.467) (25000,2.459) (25500,2.435) (26000,2.422) (26500,2.417) (27000,2.39) (27500,2.396) (28000,2.378) (28500,2.38) (29000,2.389) (29500,2.375) (30000,2.388) (30500,2.391) (31000,2.392) (31500,2.395) (32000,2.378) (32500,2.372) (33000,2.351) (33500,2.372) (34000,2.371) (34500,2.335) (35000,2.368) (35500,2.365) (36000,2.358) (36500,2.372) (37000,2.353) (37500,2.369) (38000,2.342) (38500,2.333) (39000,2.334) (39500,2.339) (40000,2.339) (40500,2.345) (41000,2.35) (41500,2.368) (42000,2.371) (42500,2.388) (43000,2.395) (43500,2.39) (44000,2.388) (44500,2.363) (45000,2.347) (45500,2.336) (46000,2.33) (46500,2.329) (47000,2.328) (47500,2.343) (48000,2.343) (48500,2.347) (49000,2.342) (49500,2.337) (50000,2.346) (50500,2.338) (51000,2.369) (51500,2.359) (52000,2.347) (52500,2.36) (53000,2.37) (53500,2.346) (54000,2.357) (54500,2.342) (55000,2.347) (55500,2.368) (56000,2.361) (56500,2.343) (57000,2.334) (57500,2.339) (58000,2.338) (58500,2.324) (59000,2.314) (59500,2.291) (60000,2.31) (60500,2.315) (61000,2.315) (61500,2.318) (62000,2.322) (62500,2.327) (63000,2.349) (63500,2.344) (64000,2.344) (64500,2.335) (65000,2.323) (65500,2.316) (66000,2.289) (66500,2.294) (67000,2.273) (67500,2.281) (68000,2.271) (68500,2.298) (69000,2.312) (69500,2.306) (70000,2.32) (70500,2.31) (71000,2.317) (71500,2.327) (72000,2.33) (72500,2.307) (73000,2.307) (73500,2.299) (74000,2.276) (74500,2.28) (75000,2.284) (75500,2.295) (76000,2.307) (76500,2.315) (77000,2.303) (77500,2.277) (78000,2.275) (78500,2.253) (79000,2.262) (79500,2.286) (80000,2.288) (80500,2.314) (81000,2.31) (81500,2.298) (82000,2.321) (82500,2.303) (83000,2.302) (83500,2.296) (84000,2.275) (84500,2.281) (85000,2.268) (85500,2.267) (86000,2.283) (86500,2.281) (87000,2.3) (87500,2.314) (88000,2.323) (88500,2.315) (89000,2.329) (89500,2.284) (90000,2.287) (90500,2.285) (91000,2.281) (91500,2.303) (92000,2.301) (92500,2.289) (93000,2.284) (93500,2.266) (94000,2.258) (94500,2.273) (95000,2.276) (95500,2.271) (96000,2.262) (96500,2.287) (97000,2.299) (97500,2.322) (98000,2.336) (98500,2.309) (99000,2.288) (99500,2.281) (100000,2.293) (100500,2.268) (101000,2.311) (101500,2.311) (102000,2.312) (102500,2.318) (103000,2.29) (103500,2.276) (104000,2.272) (104500,2.282) (105000,2.287) (105500,2.289) (106000,2.298) (106500,2.273) (107000,2.281) (107500,2.272) (108000,2.258) (108500,2.279) (109000,2.246) (109500,2.276) (110000,2.286) (110500,2.276) (111000,2.291) };
        \addplot[dashed,color=blue] coordinates { (0,2.853) (500,2.839) (1000,2.822) (1500,2.812) (2000,2.81) (2500,2.782) (3000,2.775) (3500,2.758) (4000,2.749) (4500,2.765) (5000,2.746) (5500,2.731) (6000,2.719) (6500,2.7) (7000,2.701) (7500,2.701) (8000,2.665) (8500,2.659) (9000,2.643) (9500,2.649) (10000,2.672) (10500,2.659) (11000,2.679) (11500,2.68) (12000,2.677) (12500,2.675) (13000,2.65) (13500,2.63) (14000,2.62) (14500,2.622) (15000,2.607) (15500,2.613) (16000,2.6) (16500,2.616) (17000,2.636) (17500,2.65) (18000,2.655) (18500,2.611) (19000,2.62) (19500,2.609) (20000,2.605) (20500,2.636) (21000,2.619) (21500,2.608) (22000,2.617) (22500,2.599) (23000,2.601) (23500,2.599) (24000,2.601) (24500,2.603) (25000,2.617) (25500,2.622) (26000,2.611) (26500,2.59) (27000,2.578) (27500,2.59) (28000,2.581) (28500,2.588) (29000,2.582) (29500,2.591) (30000,2.596) (30500,2.59) (31000,2.597) (31500,2.585) (32000,2.6) (32500,2.593) (33000,2.586) (33500,2.566) (34000,2.571) (34500,2.59) (35000,2.591) (35500,2.586) (36000,2.574) (36500,2.563) (37000,2.566) (37500,2.586) (38000,2.598) (38500,2.622) (39000,2.596) (39500,2.577) (40000,2.558) (40500,2.556) (41000,2.562) (41500,2.57) (42000,2.588) (42500,2.577) (43000,2.572) (43500,2.586) (44000,2.566) (44500,2.576) (45000,2.596) (45500,2.609) (46000,2.583) (46500,2.574) (47000,2.549) (47500,2.542) (48000,2.558) (48500,2.601) (49000,2.607) (49500,2.568) (50000,2.574) (50500,2.528) (51000,2.548) (51500,2.556) (52000,2.547) (52500,2.562) (53000,2.551) (53500,2.552) (54000,2.542) (54500,2.541) (55000,2.549) (55500,2.577) (56000,2.577) (56500,2.592) (57000,2.558) (57500,2.551) (58000,2.54) (58500,2.537) (59000,2.531) (59500,2.536) (60000,2.556) (60500,2.573) (61000,2.567) (61500,2.569) (62000,2.559) (62500,2.543) (63000,2.559) (63500,2.577) (64000,2.576) (64500,2.605) (65000,2.604) (65500,2.566) (66000,2.547) (66500,2.535) (67000,2.52) (67500,2.529) (68000,2.55) (68500,2.564) (69000,2.565) (69500,2.555) (70000,2.536) (70500,2.528) (71000,2.537) (71500,2.539) (72000,2.547) (72500,2.544) (73000,2.532) (73500,2.525) (74000,2.511) (74500,2.525) (75000,2.504) (75500,2.524) (76000,2.524) (76500,2.532) (77000,2.537) (77500,2.54) (78000,2.544) (78500,2.535) (79000,2.554) (79500,2.533) (80000,2.537) (80500,2.559) (81000,2.547) (81500,2.57) (82000,2.559) (82500,2.539) (83000,2.536) (83500,2.53) (84000,2.528) (84500,2.556) (85000,2.547) (85500,2.561) (86000,2.568) (86500,2.56) (87000,2.565) (87500,2.543) (88000,2.52) (88500,2.529) (89000,2.534) (89500,2.522) (90000,2.535) (90500,2.533) (91000,2.536) (91500,2.545) (92000,2.542) (92500,2.524) (93000,2.51) (93500,2.511) (94000,2.51) (94500,2.532) (95000,2.541) (95500,2.537) (96000,2.535) (96500,2.542) (97000,2.534) (97500,2.546) (98000,2.553) (98500,2.553) (99000,2.567) (99500,2.544) (100000,2.521) (100500,2.513) (101000,2.508) (101500,2.524) (102000,2.525) (102500,2.533) (103000,2.525) (103500,2.529) (104000,2.536) (104500,2.549) (105000,2.518) (105500,2.517) (106000,2.523) (106500,2.539) (107000,2.547) (107500,2.57) (108000,2.581) (108500,2.545) (109000,2.537) (109500,2.504) (110000,2.497) (110500,2.512) (111000,2.519) (111500,2.538) (112000,2.532) (112500,2.517) (113000,2.506) (113500,2.491) (114000,2.52) (114500,2.532) (115000,2.563) (115500,2.571) (116000,2.557) (116500,2.559) (117000,2.534) (117500,2.56) (118000,2.549) (118500,2.558) (119000,2.571) (119500,2.57) (120000,2.58) (120500,2.565) (121000,2.551) (121500,2.509) (122000,2.512) };
        
        \legend{w/ feedback, w/o feedback}
        \end{axis}
    \end{tikzpicture}
    \vspace{-1.5em}
    \caption{Loss curve of the GNN. \label{fig:feedback_learning_curve}}
    \end{minipage}
    \hfill
    \begin{minipage}[t]{0.48\linewidth}
    \centering
    \begin{tikzpicture}[scale=.9, every node/.style={transform shape}]
        \begin{axis}[
            width=1.1\linewidth,
            height=\linewidth,
            symbolic x coords={\OurSystem,HDP,HeteroG},
            ylabel=Time (minutes),
            xlabel=\phantom{Iter},
            ymin=0,
            bar width=.25cm,
            enlarge x limits=0.2,
            xtick=data,
        ]
            \addplot[ybar,fill=blue!60] coordinates {
                (\OurSystem,80)
                (HDP,150)
                (HeteroG,240)
            };
        \end{axis}
    \end{tikzpicture}
    \vspace{-1.5em}
    \caption{Overhead of generating a strategy on unseen device topologies.\label{fig:overhead}}
    \end{minipage}
    \vspace{-.5em}
\end{figure}

\subsection{Effect of Sufficient Factor Broadcasting\label{sec:SFB_exp}}

To reveal the potential benefits of SFB, we conduct this experiment on two machines, each equipped with one 1080Ti GPU.
We use a batch size of 4 for all models. We compare the per-iteration training time achieved with DP-NCCL and
\OurSystem, before and after enabling SFB, in Table~\ref{tab:sfb_exp}. We also include FlexFlow as a reference. When
applying SFB with DP-NCCL, we solve the SFB optimization to find a subgraph around each gradient, and replace the
gradient's AllReduce synchronization by the ``duplicate'' option. SFB brings significant speed-up in training
InceptionV3 and Transformer, suggesting that there are more low-rank structures in these models. As \OurSystem also
adopts other strategies to alleviate communication overhead, e.g., mixing PS and AllReduce, the total communication time
with \OurSystem is shorter, and the speed-up achieved by applying SFB in \OurSystem is smaller as compared to the DP
case.

We summarize the top 5 ops duplicated via SFB optimization across all 6 DNN models in Table~\ref{tab:sfb_op}. The count
indicates the total number of times when the respective ops are duplicated via SFB optimization. \OurSystem can identify
SFB opportunities beyond \texttt{MatMul}.

\begin{table}[!t]
    \caption{Top 5 operations that \OurSystem chooses to duplicate.\label{tab:sfb_op}}
    \vspace{-.5em}
    \begin{center}
    \begin{small}
    \scalebox{0.95}{
    \begin{tabular}{ |l|r| }\hline
        \textbf{Operation}    & \textbf{Count} \\\hline
        Reshape      & 341 \\\hline
        MatMul       & 336 \\\hline
        Transpose    & 89 \\\hline
        Conv2DBackpropFilter & 66 \\\hline
        Add          & 26 \\\hline
    \end{tabular}
    }
    \end{small}
    \end{center}
    \vspace{-2em}
\end{table}

\subsection{Generalizability to Unseen Device Topologies}

To evaluate generalizability of our GNN model, we randomly generate 100 unseen device topologies (in the same way as how
device topologies used for GNN training are produced, as described in Sec.~\ref{sec:exp_setup}) and use the simulator to
evaluate strategies produced by \OurSystem for training a DNN (randomly selected out of the 6 models) on each unseen
topology.

We collect the number of MCTS search iterations required for \OurSystem to find a deployment strategy that achieves
better training time than DP-NCCL. We also compare this number with the search iteration number required by pure MCTS
without using prior probabilities from the GNN, but the probabilities from a uniform distribution. In
Table~\ref{tab:generalizability_device}, we see that with prior probabilities from the GNN, \OurSystem can quickly find
strategies that outperforms DP-NCCL, while Pure MCTS needs many more iterations.

\begin{table}[!t]
    \caption{\label{tab:generalizability_device} Average number of MCTS search iterations to obtain a better strategy than DP-NCCL.}
    \vspace{-.5em}
    \begin{center}
    \begin{small}
    \scalebox{0.95}{
    \begin{tabular}{|l|r|r|}\hline
        \textbf{Model} & \textbf{Pure MCTS} & \textbf{\OurSystem} \\\hline
        InceptionV3  &      66.0 &      9.5   \\\hline
        ResNet101    &      73.4 &      4.6   \\\hline
        VGG19        &      56.6 &      17.7  \\\hline
        Transformer  &     145.0 &      121.8 \\\hline
        Bert-Small   &      97.8 &      7.9   \\\hline
    \end{tabular}
    }
    \end{small}
    \end{center}
    \vspace{-.5em}
\end{table}

\subsection{Generalizability to Unseen Computation Graphs\label{sec:eval_gen_model}}

We train \OurSystem with 5 of the models in Table~\ref{tab:models} and then produce strategies for the hold-out model
(we vary the hold-out model among the 6 DNNs). We conduct this experiments on both the testbed and the cloud. As
Table~\ref{tab:generalizability_model} shows, strategies produced for the unseen models are only marginally worse than
those for models in the training set.

\begin{table}[!t]
    \caption{\label{tab:generalizability_model} Average speed-up over DP-NCCL. \OurSystem: trained with all models. \OurSystem--: trained with other DNNs and producing strategy for one hold-out model.
    }
    \begin{center}
    \begin{small}
    \scalebox{0.95}{
    \begin{tabular}{|l|r|r|r|r|}\hline
        \multirow{2}{*}{\textbf{Model}} & \multicolumn{2}{c|}{\textbf{Testbed}} & \multicolumn{2}{c|}{\textbf{Cloud}} \\\cline{2-5}
                               & \textbf{\textit{\OurSystem}}  &  \textbf{\textit{\OurSystem--}}  & \textbf{\textit{\OurSystem}} & \textbf{\textit{\OurSystem--}} \\\hline
        InceptionV3  & 456.1\% & 456.1\% & 117.7\% & 117.7\% \\\hline
        ResNet101    &   7.7\% &   7.1\% & 12.2\% & 10.4\% \\\hline
        VGG19        & 286.2\% & 213.6\% & 43.6\% & 43.6\% \\\hline
        Transformer  &  80.3\% &  80.3\% & 15.0\% & 13.4\% \\\hline
        Bert-Small   & 298.4\% & 279.6\% & 84.5\% & 84.5\% \\\hline
    \end{tabular}
    }
    \end{small}
    \end{center}
    \vskip -2em
\end{table}

\subsection{Overhead of \OurSystem}

We compare the overhead of \OurSystem with two learning-based methods, HDP\cite{hdp} and HeteroG\cite{heterog}.
As shown in Fig.~\ref{fig:overhead}, \OurSystem is 87.5\% faster than HDP and 2x faster than HeteroG. Thanks to its
generalizability to unseen device topologies, \OurSystem only needs to run the MCTS search and GNN inference to generate
a strategy, while HeteroG requires training from scratch. HDP constantly evaluates the strategy on real
clusters during the search, which incurs a large overhead.

\section{Conclusion}\label{sec:conclusion}

This paper proposes \OurSystem, an automatic DNN deployment system that accelerates distributed training over scattered
resources. \OurSystem combines both the DNN computation graph and the device topology graph as input to a GNN, and
integrates the GNN with MCTS to identify optimized deployment strategies. With automatic partial replication and
sufficient factor broadcasting, \OurSystem can better utilize heterogeneous resources and reduce communication overhead
for DNN training. In our experiments, \OurSystem achieves up to 4.56x speed-up as compared to representative existing
schemes. \OurSystem is generic and efficient in producing good strategies for both unseen device topologies and DNN
models without re-training. 

As a future direction, we plan to extend \OurSystem to support pipeline parallelism, by expanding the replication plan
to include a ``pipeline'' option. The graph compiler would need to add appropriate control dependencies in the
distributed graph for ops to process different micro-batches in the same pipeline stage, to achieve efficient
pipelining.

\section*{Acknowledgement}

This work was supported in part by Alibaba Group through Alibaba Innovative Research (AIR) Program, and Hong Kong RGC under the contracts HKU 17204619 and 17208920.

\ifCLASSOPTIONcaptionsoff
  \newpage
\fi

\bibliography{main}

\begin{thebibliography}{10}
\providecommand{\url}[1]{#1}
\csname url@samestyle\endcsname
\providecommand{\newblock}{\relax}
\providecommand{\bibinfo}[2]{#2}
\providecommand{\BIBentrySTDinterwordspacing}{\spaceskip=0pt\relax}
\providecommand{\BIBentryALTinterwordstretchfactor}{4}
\providecommand{\BIBentryALTinterwordspacing}{\spaceskip=\fontdimen2\font plus
\BIBentryALTinterwordstretchfactor\fontdimen3\font minus
  \fontdimen4\font\relax}
\providecommand{\BIBforeignlanguage}[2]{{%
\expandafter\ifx\csname l@#1\endcsname\relax
\typeout{** WARNING: IEEEtran.bst: No hyphenation pattern has been}%
\typeout{** loaded for the language `#1'. Using the pattern for}%
\typeout{** the default language instead.}%
\else
\language=\csname l@#1\endcsname
\fi
#2}}
\providecommand{\BIBdecl}{\relax}
\BIBdecl

\bibitem{resnet}
K.~He, X.~Zhang, S.~Ren, and J.~Sun, ``Deep residual learning for image
  recognition,'' in \emph{Proceedings of the IEEE conference on computer vision
  and pattern recognition}, 2016, pp. 770--778.

\bibitem{inception}
C.~Szegedy, V.~Vanhoucke, S.~Ioffe, J.~Shlens, and Z.~Wojna, ``Rethinking the
  inception architecture for computer vision,'' in \emph{Proceedings of the
  IEEE conference on computer vision and pattern recognition}, 2016, pp.
  2818--2826.

\bibitem{gpt3}
T.~B. Brown, B.~Mann, N.~Ryder, M.~Subbiah, J.~Kaplan, P.~Dhariwal,
  A.~Neelakantan, P.~Shyam, G.~Sastry, A.~Askell \emph{et~al.}, ``Language
  models are few-shot learners,'' \emph{arXiv preprint arXiv:2005.14165}, 2020.

\bibitem{palm}
A.~Chowdhery, S.~Narang, J.~Devlin, M.~Bosma, G.~Mishra, A.~Roberts, P.~Barham,
  H.~W. Chung, C.~Sutton, S.~Gehrmann \emph{et~al.}, ``Palm: Scaling language
  modeling with pathways,'' \emph{arXiv preprint arXiv:2204.02311}, 2022.

\bibitem{wang2018billion}
J.~Wang, P.~Huang, H.~Zhao, Z.~Zhang, B.~Zhao, and D.~L. Lee, ``Billion-scale
  commodity embedding for e-commerce recommendation in alibaba,'' in
  \emph{Proceedings of the 24th ACM SIGKDD International Conference on
  Knowledge Discovery \& Data Mining}, 2018, pp. 839--848.

\bibitem{bert}
J.~Devlin, M.-W. Chang, K.~Lee, and K.~Toutanova, ``Bert: Pre-training of deep
  bidirectional transformers for language understanding,'' \emph{arXiv preprint
  arXiv:1810.04805}, 2018.

\bibitem{biobert}
J.~Lee, W.~Yoon, S.~Kim, D.~Kim, S.~Kim, C.~H. So, and J.~Kang, ``Biobert: a
  pre-trained biomedical language representation model for biomedical text
  mining,'' \emph{Bioinformatics}, vol.~36, no.~4, pp. 1234--1240, 2020.

\bibitem{placeto}
R.~Addanki, S.~B. Venkatakrishnan, S.~Gupta, H.~Mao, and M.~Alizadeh,
  ``Placeto: Learning generalizable device placement algorithms for distributed
  machine learning,'' \emph{arXiv preprint arXiv:1906.08879}, 2019.

\bibitem{gdp}
Y.~Zhou, S.~Roy, A.~Abdolrashidi, D.~Wong, P.~C. Ma, Q.~Xu, M.~Zhong, H.~Liu,
  A.~Goldie, A.~Mirhoseini \emph{et~al.}, ``Gdp: Generalized device placement
  for dataflow graphs,'' \emph{arXiv preprint arXiv:1910.01578}, 2019.

\bibitem{heterog}
X.~Yi, S.~Zhang, Z.~Luo, G.~Long, L.~Diao, C.~Wu, Z.~Zheng, J.~Yang, and
  W.~Lin, ``Optimizing distributed training deployment in heterogeneous gpu
  clusters,'' in \emph{Proceedings of the 16th International Conference on
  emerging Networking EXperiments and Technologies}, 2020, pp. 93--107.

\bibitem{autosync}
H.~Zhang, Y.~Li, Z.~Deng, X.~Liang, L.~Carin, and E.~Xing, ``Autosync: Learning
  to synchronize for data-parallel distributed deep learning,'' \emph{Advances
  in Neural Information Processing Systems}, vol.~33, 2020.

\bibitem{horovod}
A.~Sergeev and M.~Del~Balso, ``Horovod: fast and easy distributed deep learning
  in tensorflow,'' \emph{arXiv preprint arXiv:1802.05799}, 2018.

\bibitem{nccl}
S.~Jeaugey, ``Nccl 2.0,'' in \emph{GPU Technology Conference (GTC)}, 2017.

\bibitem{ps}
M.~Li, D.~G. Andersen, J.~W. Park, A.~J. Smola, A.~Ahmed, V.~Josifovski,
  J.~Long, E.~J. Shekita, and B.-Y. Su, ``Scaling distributed machine learning
  with the parameter server,'' in \emph{11th $\{$USENIX$\}$ Symposium on
  Operating Systems Design and Implementation ($\{$OSDI$\}$ 14)}, 2014, pp.
  583--598.

\bibitem{byteps}
Y.~Jiang, Y.~Zhu, C.~Lan, B.~Yi, Y.~Cui, and C.~Guo, ``A unified architecture
  for accelerating distributed $\{$DNN$\}$ training in heterogeneous gpu/cpu
  clusters,'' in \emph{14th $\{$USENIX$\}$ Symposium on Operating Systems
  Design and Implementation ($\{$OSDI$\}$ 20)}, 2020, pp. 463--479.

\bibitem{dgc}
Y.~Lin, S.~Han, H.~Mao, Y.~Wang, and W.~J. Dally, ``Deep gradient compression:
  Reducing the communication bandwidth for distributed training,'' \emph{arXiv
  preprint arXiv:1712.01887}, 2017.

\bibitem{parallax}
S.~Kim, G.-I. Yu, H.~Park, S.~Cho, E.~Jeong, H.~Ha, S.~Lee, J.~S. Jeong, and
  B.-G. Chun, ``Parallax: Sparsity-aware data parallel training of deep neural
  networks,'' \emph{arXiv preprint arXiv:1808.02621}, 2018.

\bibitem{hdp}
A.~Mirhoseini, A.~Goldie, H.~Pham, B.~Steiner, Q.~V. Le, and J.~Dean, ``A
  hierarchical model for device placement,'' in \emph{International Conference
  on Learning Representations}, 2018.

\bibitem{grl}
A.~Mirhoseini, H.~Pham, Q.~V. Le, B.~Steiner, R.~Larsen, Y.~Zhou, N.~Kumar,
  M.~Norouzi, S.~Bengio, and J.~Dean, ``Device placement optimization with
  reinforcement learning,'' in \emph{International Conference on Machine
  Learning}.\hskip 1em plus 0.5em minus 0.4em\relax PMLR, 2017, pp. 2430--2439.

\bibitem{heterognn}
C.~Zhang, D.~Song, C.~Huang, A.~Swami, and N.~V. Chawla, ``Heterogeneous graph
  neural network,'' in \emph{Proceedings of the 25th ACM SIGKDD International
  Conference on Knowledge Discovery \& Data Mining}, 2019, pp. 793--803.

\bibitem{mcts}
L.~Kocsis and C.~Szepesv{\'a}ri, ``Bandit based monte-carlo planning,'' in
  \emph{European conference on machine learning}.\hskip 1em plus 0.5em minus
  0.4em\relax Springer, 2006, pp. 282--293.

\bibitem{puct}
D.~Auger, A.~Couetoux, and O.~Teytaud, ``Continuous upper confidence trees with
  polynomial exploration--consistency,'' in \emph{Joint European Conference on
  Machine Learning and Knowledge Discovery in Databases}.\hskip 1em plus 0.5em
  minus 0.4em\relax Springer, 2013, pp. 194--209.

\bibitem{sfb}
P.~Xie, J.~K. Kim, Y.~Zhou, Q.~Ho, A.~Kumar, Y.~Yu, and E.~Xing, ``Distributed
  machine learning via sufficient factor broadcasting,'' \emph{arXiv preprint
  arXiv:1511.08486}, 2015.

\bibitem{tensorflow}
M.~Abadi, P.~Barham, J.~Chen, Z.~Chen, A.~Davis, J.~Dean, M.~Devin,
  S.~Ghemawat, G.~Irving, M.~Isard \emph{et~al.}, ``Tensorflow: A system for
  large-scale machine learning,'' in \emph{12th $\{$USENIX$\}$ symposium on
  operating systems design and implementation ($\{$OSDI$\}$ 16)}, 2016.

\bibitem{pytorch}
A.~Paszke, S.~Gross, S.~Chintala, G.~Chanan, E.~Yang, Z.~DeVito, Z.~Lin,
  A.~Desmaison, L.~Antiga, and A.~Lerer, ``Automatic differentiation in
  pytorch,'' 2017.

\bibitem{mxnet}
T.~Chen, M.~Li, Y.~Li, M.~Lin, N.~Wang, M.~Wang, T.~Xiao, B.~Xu, C.~Zhang, and
  Z.~Zhang, ``Mxnet: A flexible and efficient machine learning library for
  heterogeneous distributed systems,'' \emph{arXiv preprint arXiv:1512.01274},
  2015.

\bibitem{mp}
S.~Pal, E.~Ebrahimi, A.~Zulfiqar, Y.~Fu, V.~Zhang, S.~Migacz, D.~Nellans, and
  P.~Gupta, ``Optimizing multi-gpu parallelization strategies for deep learning
  training,'' \emph{IEEE Micro}, vol.~39, no.~5, pp. 91--101, 2019.

\bibitem{go}
Y.~Zhou, S.~Roy, A.~Abdolrashidi, D.~Wong, P.~Ma, Q.~Xu, H.~Liu, M.~P.
  Phothilimtha, S.~Wang, A.~Goldie \emph{et~al.}, ``Transferable graph
  optimizers for ml compilers,'' \emph{arXiv preprint arXiv:2010.12438}, 2020.

\bibitem{spotlight}
Y.~Gao, L.~Chen, and B.~Li, ``Spotlight: Optimizing device placement for
  training deep neural networks,'' in \emph{International Conference on Machine
  Learning}.\hskip 1em plus 0.5em minus 0.4em\relax PMLR, 2018, pp. 1676--1684.

\bibitem{flexflow}
Z.~Jia, M.~Zaharia, and A.~Aiken, ``Beyond data and model parallelism for deep
  neural networks.'' \emph{Proceedings of Machine Learning and Systems},
  vol.~1, pp. 1--13, 2019.

\bibitem{regal}
A.~Paliwal, F.~Gimeno, V.~Nair, Y.~Li, M.~Lubin, P.~Kohli, and O.~Vinyals,
  ``Reinforced genetic algorithm learning for optimizing computation graphs,''
  \emph{arXiv preprint arXiv:1905.02494}, 2019.

\bibitem{pesto}
U.~U. Hafeez, X.~Sun, A.~Gandhi, and Z.~Liu, ``Towards optimal placement and
  scheduling of dnn operations with pesto,'' in \emph{Proceedings of the 22nd
  International Middleware Conference}, 2021, pp. 39--51.

\bibitem{hived}
H.~Zhao, Z.~Han, Z.~Yang, Q.~Zhang, F.~Yang, L.~Zhou, M.~Yang, F.~C. Lau,
  Y.~Wang, Y.~Xiong \emph{et~al.}, ``Hived: Sharing a $\{$GPU$\}$ cluster for
  deep learning with guarantees,'' in \emph{14th $\{$USENIX$\}$ Symposium on
  Operating Systems Design and Implementation ($\{$OSDI$\}$ 20)}, 2020, pp.
  515--532.

\bibitem{mlaas}
Q.~Weng, W.~Xiao, Y.~Yu, W.~Wang, C.~Wang, J.~He, Y.~Li, L.~Zhang, W.~Lin, and
  Y.~Ding, ``{MLaaS} in the wild: Workload analysis and scheduling in
  {Large-Scale} heterogeneous {GPU} clusters,'' in \emph{19th USENIX Symposium
  on Networked Systems Design and Implementation (NSDI 22)}, 2022, pp.
  945--960.

\bibitem{chilimbi2014project}
T.~Chilimbi, Y.~Suzue, J.~Apacible, and K.~Kalyanaraman, ``Project adam:
  Building an efficient and scalable deep learning training system,'' in
  \emph{11th $\{$USENIX$\}$ Symposium on Operating Systems Design and
  Implementation ($\{$OSDI$\}$ 14)}, 2014, pp. 571--582.

\bibitem{zhang2017poseidon}
H.~Zhang, Z.~Zheng, S.~Xu, W.~Dai, Q.~Ho, X.~Liang, Z.~Hu, J.~Wei, P.~Xie, and
  E.~P. Xing, ``Poseidon: An efficient communication architecture for
  distributed deep learning on $\{$GPU$\}$ clusters,'' in \emph{2017
  $\{$USENIX$\}$ Annual Technical Conference ($\{$USENIX$\}$$\{$ATC$\}$ 17)},
  2017, pp. 181--193.

\bibitem{quantization1}
F.~Seide, H.~Fu, J.~Droppo, G.~Li, and D.~Yu, ``1-bit stochastic gradient
  descent and its application to data-parallel distributed training of speech
  dnns,'' in \emph{Fifteenth Annual Conference of the International Speech
  Communication Association}, 2014.

\bibitem{sparsification1}
N.~Strom, ``Scalable distributed dnn training using commodity gpu cloud
  computing,'' in \emph{Sixteenth Annual Conference of the International Speech
  Communication Association}, 2015.

\bibitem{confuciux}
S.-C. Kao, G.~Jeong, and T.~Krishna, ``Confuciux: Autonomous hardware resource
  assignment for dnn accelerators using reinforcement learning,'' in \emph{2020
  53rd Annual IEEE/ACM International Symposium on Microarchitecture
  (MICRO)}.\hskip 1em plus 0.5em minus 0.4em\relax IEEE, 2020, pp. 622--636.

\bibitem{rloverfit1}
C.~Zhang, O.~Vinyals, R.~Munos, and S.~Bengio, ``A study on overfitting in deep
  reinforcement learning,'' \emph{arXiv preprint arXiv:1804.06893}, 2018.

\bibitem{rloverfit2}
X.~Song, Y.~Jiang, S.~Tu, Y.~Du, and B.~Neyshabur, ``Observational overfitting
  in reinforcement learning,'' \emph{arXiv preprint arXiv:1912.02975}, 2019.

\bibitem{tfslim}
{Sergio Guadarrama, Nathan Silberman}, ``{TensorFlow-Slim}: A lightweight
  library for defining, training and evaluating complex models in tensorflow,''
  \url{https://github.com/google-research/tf-slim}, 2016.

\bibitem{metis}
G.~Karypis and V.~Kumar, ``A fast and high quality multilevel scheme for
  partitioning irregular graphs,'' \emph{SIAM Journal on scientific Computing},
  vol.~20, no.~1, pp. 359--392, 1998.

\bibitem{bao2020preemptive}
Y.~Bao, Y.~Peng, Y.~Chen, and C.~Wu, ``Preemptive all-reduce scheduling for
  expediting distributed dnn training,'' in \emph{IEEE INFOCOM 2020-IEEE
  Conference on Computer Communications}.\hskip 1em plus 0.5em minus
  0.4em\relax IEEE, 2020.

\bibitem{oversmoothing1}
D.~Chen, Y.~Lin, W.~Li, P.~Li, J.~Zhou, and X.~Sun, ``Measuring and relieving
  the over-smoothing problem for graph neural networks from the topological
  view,'' in \emph{Proceedings of the AAAI Conference on Artificial
  Intelligence}, vol.~34, no.~04, 2020, pp. 3438--3445.

\bibitem{oversmoothing2}
C.~Cai and Y.~Wang, ``A note on over-smoothing for graph neural networks,''
  \emph{arXiv preprint arXiv:2006.13318}, 2020.

\bibitem{gat}
P.~Veli{\v{c}}kovi{\'c}, G.~Cucurull, A.~Casanova, A.~Romero, P.~Lio, and
  Y.~Bengio, ``Graph attention networks,'' \emph{arXiv preprint
  arXiv:1710.10903}, 2017.

\bibitem{cbc}
J.~Forrest, T.~Ralphs, S.~Vigerske, L.~Hafer, B.~Kristjansson, J.~Fasano,
  E.~Straver, M.~Lubin, H.~Santos, R.~Lougee \emph{et~al.}, ``coin-or/cbc:
  Version 2.9. 9,'' \emph{URL http://dx.doi.org/10.5281/zenodo}, vol. 1317566,
  2018.

\bibitem{dgl}
M.~Wang, D.~Zheng, Z.~Ye, Q.~Gan, M.~Li, X.~Song, J.~Zhou, C.~Ma, L.~Yu,
  Y.~Gai, T.~Xiao, T.~He, G.~Karypis, J.~Li, and Z.~Zhang, ``Deep graph
  library: A graph-centric, highly-performant package for graph neural
  networks,'' \emph{arXiv preprint arXiv:1909.01315}, 2019.

\bibitem{adam}
D.~P. Kingma and J.~Ba, ``Adam: A method for stochastic optimization,''
  \emph{arXiv preprint arXiv:1412.6980}, 2014.

\bibitem{vgg}
K.~Simonyan and A.~Zisserman, ``Very deep convolutional networks for
  large-scale image recognition,'' \emph{arXiv preprint arXiv:1409.1556}, 2014.

\bibitem{transformer}
A.~Vaswani, N.~Shazeer, N.~Parmar, J.~Uszkoreit, L.~Jones, A.~N. Gomez,
  L.~Kaiser, and I.~Polosukhin, ``Attention is all you need,'' \emph{arXiv
  preprint arXiv:1706.03762}, 2017.

\bibitem{post}
Y.~Gao, L.~Chen, and B.~Li, ``Post: Device placement with cross-entropy
  minimization and proximal policy optimization,'' \emph{Advances in Neural
  Information Processing Systems}, vol.~31, 2018.

\bibitem{baechi}
B.~Jeon, L.~Cai, P.~Srivastava, J.~Jiang, X.~Ke, Y.~Meng, C.~Xie, and I.~Gupta,
  ``Baechi: fast device placement of machine learning graphs,'' in
  \emph{Proceedings of the 11th ACM Symposium on Cloud Computing}, 2020, pp.
  416--430.

\bibitem{legion}
M.~Bauer, S.~Treichler, E.~Slaughter, and A.~Aiken, ``Legion: Expressing
  locality and independence with logical regions,'' in \emph{SC'12: Proceedings
  of the International Conference on High Performance Computing, Networking,
  Storage and Analysis}.\hskip 1em plus 0.5em minus 0.4em\relax IEEE, 2012, pp.
  1--11.

\end{thebibliography}
\bibliographystyle{IEEEtran}

\end{document}